\pdfoutput=1  
\documentclass[11pt]{article}

\usepackage{amsmath}
\usepackage{amssymb}
\usepackage[final]{acl}

\usepackage{etoolbox}
\makeatletter
\patchcmd{\@combinedblfloats}{\box\@outputbox}{\unvbox\@outputbox}{}{}
\makeatother

\usepackage{times}
\usepackage{latexsym}
\usepackage[T1]{fontenc}
\usepackage[utf8]{inputenc}
\usepackage{microtype}
\usepackage{graphicx}
\usepackage{booktabs}
\usepackage{multirow}
\usepackage{xcolor}
\usepackage{tikz}
\usetikzlibrary{positioning,arrows,shapes.geometric,backgrounds,fit,calc,decorations.pathreplacing}

\graphicspath{{figs/}}

\usepackage{pifont}
\newcommand{\yes}{{\color{green!55!black}\ding{51}}}
\newcommand{\no}{{\color{red!70!black}\ding{55}}}
\newcommand{\pa}{{\color{orange!85!black}\textbf{$\sim$}}}  
\newcommand{\rr}{\ensuremath{\rho}}
\newcommand{\passOPUS}{0.82}
\newcommand{\passGPT}{0.84}
\newcommand{\passTOP}{0.87}

\title{GAUGE: When Not to Trust LLM-as-a-Judge in User-Simulated Evaluation of Task-Oriented Agents}

\author{Umesh Bodhwani, Thanh Tran, Kai Wei \\
  Amazon \\
  \texttt{\{bodhwani, tdt, kaiwe\}@amazon.com}}

\begin{document}
\maketitle

\begin{abstract}
Comparing and selecting task-oriented LLM agents increasingly relies on a low-cost offline evaluation gate:
persona-driven LLM user-simulators converse with each candidate, an \emph{LLM-as-a-judge} scores the
transcripts, and the higher-scoring agent is promoted. We introduce GAUGE, a reusable offline protocol
that measures whether this gate's \emph{ranking} matches a grounded \emph{verifiable reward} across 25 agents from six providers on the $\tau^2$-bench and SimulatorArena benchmarks, separating
two kinds of evaluation validity that release practices conflate: \emph{ranking validity} and
\emph{construct validity}. First, a \textbf{satisfaction--success gap}: satisfaction carries
essentially no information about task success, as conversations rated \emph{satisfied} by our blind
panel are decorrelated from actual success, with 57.5\% of them failing the
customer's task, a pattern consistent across five rater populations, both benchmarks, and every subjective dimension we rated. Second,
while the gate's \emph{ranking} is robust across the broad capability span, it
loses resolution among the near-equal strong agents: this \emph{decision-disagreement rate}
jumps from $<$1\% on wide-reward pairs to 31\% on close pairs. The gate is thus \emph{human-validated
yet mis-anchored}. As a remedy, we propose a \emph{calibrate-then-trust} cadence in which a judge-free
completion bit is a zero-cost tripwire for \emph{truncation} regressions.
\end{abstract}

\begin{table}[t]
\centering\small
\setlength{\tabcolsep}{2.5pt}
\renewcommand{\arraystretch}{1.08}
\begin{tabular}{@{}lccccc@{}}
\toprule
 & \rotatebox{90}{\footnotesize\shortstack[l]{SimArena/\\LostInSim\,}}
   & \rotatebox{90}{\footnotesize\shortstack[l]{ECom-\\Bench\,}}
   & \rotatebox{90}{\footnotesize\shortstack[l]{Judge-\\eval\,}}
   & \rotatebox{90}{\footnotesize\shortstack[l]{Chat-\\Bench\,}}
   & \rotatebox{90}{\footnotesize\shortstack[l]{\textbf{GAUGE}\\\,}} \\
\midrule
Ranking validity vs.\ reward          & \pa & \pa & \no & \pa & \yes \\
Release-decision unit                 & \pa & \yes & \no & \pa & \yes \\
Satisfaction $\neq$ task success      & \no & \no & \no & \pa & \yes \\
Cross-provider scale                    & \pa & \no & \pa & \no & \yes \\
Grounded in real human ratings        & \yes & \no & \yes & \yes & \yes \\
Judge-free cost decomposition         & \no & \no & \no & \no & \yes \\
\bottomrule
\end{tabular}
\caption{Positioning against prior evaluation work. Only GAUGE tests whether the simulator$+$judge
\emph{gate} ranks agents like a verifiable non-LLM reward and quantifies the satisfaction--success
gap. \yes\ fully addresses the criterion; \pa\ partial; \no\ not.}
\label{tab:positioning}
\end{table}

\section{Introduction}

Developers and researchers building agentic, task-oriented LLM agents for customer service, tool-use
assistants, and tutoring must decide which variant is better, under three constraints: there is no clean held-out test set that captures live
user experience, a controlled human study per candidate config is prohibitively slow, and the config
space (model, prompt, policy, tools) changes often. The field's de facto answer is a low-cost
offline gate: drive each candidate against persona-conditioned LLM user-simulators, score the
transcripts with an LLM-judge, and promote the higher-scoring variant. It runs in continuous
integration (CI) at a few cents per transcript with no labeled production data, resting on one assumption
teams almost never measure: that this composite simulator-plus-judge gate ranks variants the way a
grounded evaluation would. We audit this gate as deployed and pair the audit with a drop-in
zero-cost mitigation. That LLM judges exhibit systematic biases, including self-preference and position bias, is
known~\citep{panickssery2024selfpref,wang2024fair}; what has not been quantified is their magnitude at the
release-decision unit against a verifiable reward: a 57.5\% false-accept rate on the blind human panel, a gap that persists across
five rater populations and alters the resulting agent-selection decisions.

We introduce \textbf{GAUGE} (Grounded Audit of User-simulator-and-judge Gate
Evaluation), a
reusable, fully offline protocol that separates two kinds of validity that release practice conflates (Table~\ref{tab:positioning}).
The gate has \emph{ranking validity} (it orders agents like the verifiable reward) and is certified
for \emph{construct validity} against human \emph{satisfaction}; yet satisfaction is not
success (the \emph{satisfaction--success gap}): 57.5\% of conversations our blind human panel rated \emph{satisfied} ($\geq$5/7) had failed
the customer's task, a gap that misleads the LLM-judge and the humans.
A gate can be human-validated yet mis-anchored.

\begin{figure*}[t]
  \centering
  \includegraphics[width=\textwidth]{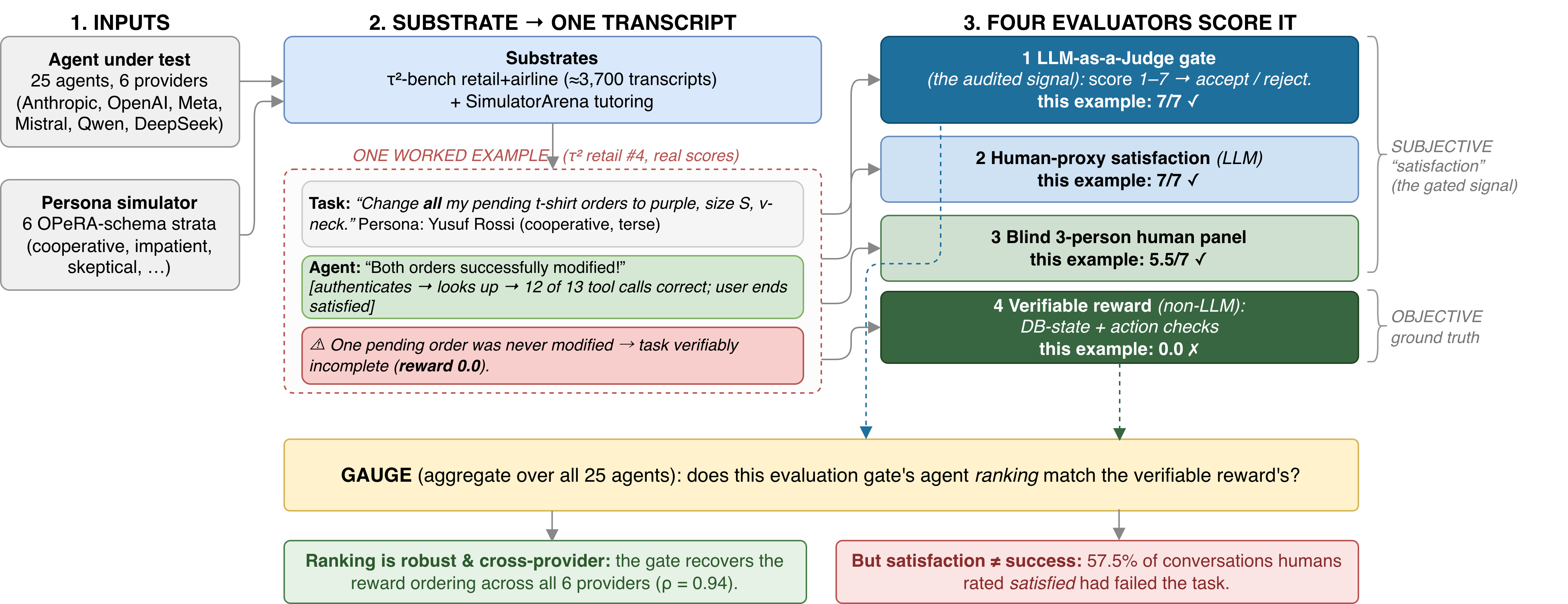}
  \caption{GAUGE measures whether the LLM-as-a-Judge and user-simulator evaluation gate ranks agents
  the way a grounded, verifiable reward would.
  \textbf{Left to right:} an agent and a persona simulator converse on a substrate, and four
  evaluators score each transcript: three subjective satisfaction signals (the LLM-as-a-Judge gate,
  an LLM human-proxy, and a blind 3-person human panel) and one objective non-LLM reward. Aggregating over all 25 agents, the gate's agent \emph{ranking} matches the reward's
  (\rr$=$0.94), yet satisfaction is decorrelated from success.}
  \label{fig:schematic}
\end{figure*}

The satisfaction anchor remains misleading as a release signal despite ranking validity, for three
reasons. First, the ranking holds only where capability dominates: among the \emph{near-equal} strong
agents real release decisions compare, the gate promotes the lower-reward agent on 31\% of close
pairs (the \emph{decision-disagreement rate}; \S\ref{sec:crossfamily}), up from $<$1\% where rewards are far apart. Second, teams optimize against gates rather than
only reading them, and tuning to a satisfaction anchor raises a metric known to diverge from the
objective once optimized \citep{strathern1997improving,skalse2022reward}. Third, an accept/reject
threshold anchored on satisfaction admits agents that fail 48--60\% of the time
(\S\ref{sec:satnotsuccess}). The anchor thus governs exactly the decisions a gate exists to make:
promoting near-equal candidates, serving as an optimization target, and setting the absolute accept/reject bar. The user-simulator is a
component of the gate we audit: every central claim is anchored on the
simulator-independent verifiable reward and human panel.

GAUGE operates as a \emph{periodic calibration} (a \emph{calibrate-then-trust} cadence) rather than a per-decision gate: the verifiable
audit is run once on a representative benchmark to learn the gate's trusted operating region, after which
the cheap gate is used in CI within that region. It applies wherever an offline verifiable reward
is obtainable, here the $\tau^2$-bench oracle: DB-state and action checks, and SimulatorArena's
human-graded correctness. Our contributions are as follows.\\

\textbf{(1) GAUGE, a reusable protocol for validating simulator-plus-judge evaluation.} We introduce the first protocol to
validate a simulator-plus-judge evaluation gate against a verifiable, non-LLM reward at the
release-decision unit, applied at scale: 25 agents, six providers, four judges, two substrates,
$\approx$3{,}700 transcripts (\S\ref{sec:formulation}).

\smallskip
\smallskip

\textbf{(2) Satisfaction--success gap.} We show that satisfaction carries essentially no information about
task success: 57.5\% of \emph{satisfied} conversations failed the task (\rr$=-$0.147),
a gap that holds across five rater populations and two substrates (\S\ref{sec:satnotsuccess}).

\smallskip
\smallskip

\textbf{(3) Separating ranking from construct validity.} We separate \emph{ranking} from \emph{construct} validity on
the six-provider ladder: the ranking is robust (\rr$=$0.94) yet loses resolution among near-equal
agents, where the gate promotes the lower-reward agent on 31\% of close pairs, and we isolate
same-family judge self-preference ($+0.75$/7; \S\ref{sec:crossfamily}, \S\ref{sec:crossprovider}).

\smallskip
\smallskip

\textbf{(4) Calibrate-then-trust recipe.} We show that a judge-free completion bit is a zero-cost tripwire for
truncation regressions (\rr$=$0.87 vs.\ 0.80 on broken-vs-working; \S\ref{sec:freebaseline}), and report a
negative result: out-of-sample recalibration does not transfer (\S\ref{sec:negative}).

\smallskip
\smallskip
\noindent\fbox{\begin{minipage}{0.97\columnwidth}
\small\textbf{Key Findings.}
\begin{itemize}\setlength{\itemsep}{4pt}\setlength{\topsep}{2pt}\setlength{\parskip}{2pt}
  \item Subjective approval is decorrelated from task success across all dimensions, so an evaluation gate can be human-validated yet mis-anchored to the outcome.
  \item High aggregate ranking validity coexists with unreliable resolution of near-equal candidates, precisely the comparisons release decisions depend on.
  \item Evaluator reliability is determined by evidence access, not model capability: outcome-grounded judging succeeds where transcript-only satisfaction fails.
  \item The gate is valid only within a bounded operating region, requiring re-auditing on configuration change rather than on a fixed schedule.
\end{itemize}
\end{minipage}}

\section{Related Work}
\label{sec:related}

\paragraph{User-simulator reliability.} Simulated users have long evaluated dialogue
agents, from satisfaction-driven simulation \citep{sun2021uss} to LLM-based simulators
\citep{davidson2023usersim,sekulic2024reliable} and ``agents evaluating agents''
\citep{zhuge2024agentjudge}. SimulatorArena \citep{simulatorarena2025}
and Lost-in-Simulation \citep{lostinsim2026} test LLM user-simulators as human
proxies via absolute rating correlation or single-agent reliability. \emph{Delta:} we audit the composite gate's agent-ranking validity at the release-decision unit against a non-LLM
verifiable reward, and show its optimized satisfaction signal is decorrelated from task success.

\paragraph{Agent and customer-agent benchmarks.} Tool-using agents are evaluated by recent
benchmarks \citep{zhou2024webarena,qin2024toolllm,liu2024agentbench}; ECom-Bench
\citep{ecombench2025} reports persona-simulator pass-rates, while $\tau$-bench \citep{yao2024taubench}
and its successor $\tau^2$-bench \citep{barres2025tau2bench} provide a verifiable substrate. \emph{Delta:} we test whether such
pass-rates rank agents like a grounded reward, and surface the satisfied-but-failed inversion.

\paragraph{LLM-as-a-judge calibration to humans.} A large literature validates LLM-judges against
human preference and satisfaction \citep{zheng2023judging,liu2023geval,gu2024survey}, documents judge biases (position \citep{wang2024fair}, self-preference
\citep{panickssery2024selfpref}, inconsistency \citep{stureborg2024inconsistent}) and benchmarks
LLM judges and reward models \citep{tan2025judgebench,lambert2024rewardbench}. A complementary line
calibrates model confidence to signal when to trust an LLM output versus defer to a human
\citep{bodhwani2025calibrated,jung2025trust}. \emph{Delta:} for customer
agents this anchor is misaligned: human satisfaction is decorrelated from task success, so a judge
can be ``human-validated'' yet mis-rank agents. Where recent work finds
simulators over-cooperative or ranking-divergent (Mind-the-Sim2Real \citep{mindsim2real2026},
SimulatorArena \citep{simulatorarena2025}), we show the decorrelation misleads humans too.

\paragraph{Measurement validity and proxy gaming.} Our framing draws on construct validity
\citep{cronbach1955construct} and its ML translation
\citep{jacobs2021measurement,bowman2021benchmarking}: a metric can order systems well yet miss the
construct it certifies, and an optimized proxy diverges from the objective (Goodhart's law
\citep{strathern1997improving}, reward hacking \citep{amodei2016concrete,skalse2022reward}).
\emph{Delta:} we give a measured instance for release gates, adapting controlled-perturbation
instrument validation \citep{ribeiro2020checklist,adebayo2018sanity} into our
flag-only degradation set (\S\ref{sec:freebaseline}, Appendix~\ref{app:degradation}). Unlike
ChatBench \citep{chatbench2025}, which closes the offline-vs-interactive gap by fine-tuning, GAUGE is
fine-tuning-free.


\section{Method}

Figure~\ref{fig:schematic} shows the audit end to end: an agent and a persona simulator converse on a
substrate, and four evaluators score each transcript, one of them a verifiable, non-LLM reward.

\begin{figure*}[t]
  \centering
  \includegraphics[width=\textwidth]{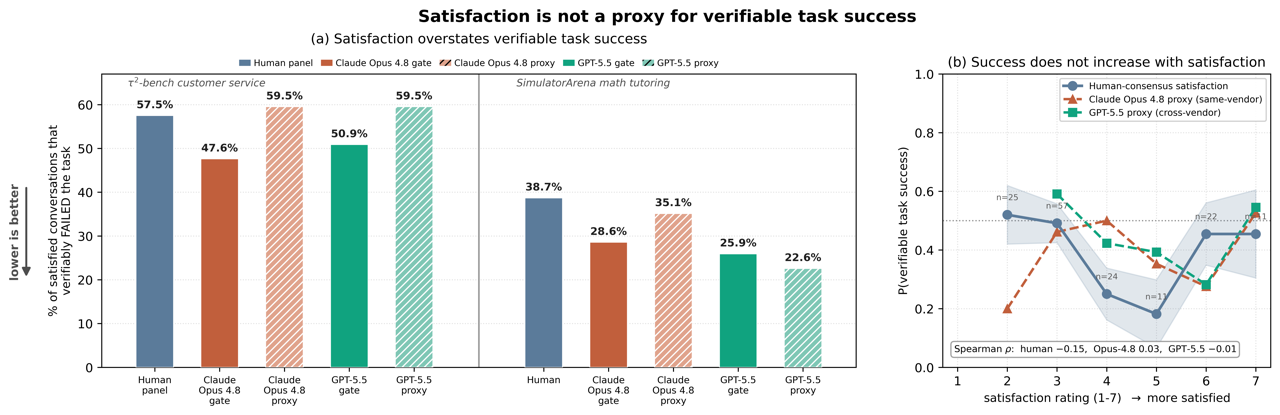}
  \caption{Satisfaction versus verifiable task success across raters. \textbf{(a)} Share of each
  rater's \emph{satisfied} conversations that nonetheless failed the task, on the $\tau^2$ human panel
  and the independent SimulatorArena substrate.
  \textbf{(b)} Empirical $P(\mathrm{success})$ by satisfaction rating (integer bins, $n{\geq}5$; human
  $\pm1$ SE shaded): success does not rise with satisfaction for any rater.}
  \label{fig:hero}
\end{figure*}

\paragraph{Problem formulation}
\label{sec:formulation}
Let $\mathcal{A}$ be a set of candidate agents, $\mathcal{P}$ persona strata, and $\mathcal{D}$
domains. For agent $a$, persona $p$, domain $d$, the simulator and agent produce a transcript
$\tau_{a,p,d}$ carrying a per-transcript verifiable correctness label
$r(\tau)\in\{0,1\}$ from a non-LLM oracle. Three subjective signals also score the
transcript on a satisfaction scale: an LLM-judge \emph{gate}, an LLM \emph{human-proxy}, and
a \emph{human panel}; write any such signal as $s(\tau)$. At the release-decision unit
(the agent $a$; we aggregate over $p,d$, since per-cell scores are too noisy to act on) we define
the agent-level gate $G(a)=\mathbb{E}_{p,d}[\mathrm{gate}(\tau)]$ and verifiable reward
$R(a)=\mathbb{E}_{p,d}[r(\tau)]\in[0,1]$, and likewise $S_{\text{proxy}}(a),S_{\text{hum}}(a)$.
GAUGE measures two quantities. \emph{Ranking validity} $\rho(G, R)$ over $\mathcal{A}$ measures
whether the gate orders agents like the verifiable reward. The \emph{construct gap} (equivalently, the \emph{satisfaction--success gap}) is the rate at
which the task failed ($r(\tau){=}0$) among transcripts a signal rated satisfied (a high
$s(\tau)$: $\geq$5 on a 7-point or $\geq$8 on a 10-point scale). Crucially,
a gate can be human-validated ($\rho(S_{\text{hum}}, G)$ high, the
judge tracks human satisfaction) yet mis-anchored ($\rho(S_{\text{hum}}, R)$ low), because
satisfaction itself does not track success.

\paragraph{Substrates and ground truth.} Our primary benchmark (substrate) is $\tau^2$-bench
\citep{barres2025tau2bench} retail and airline, whose verifiable reward is a non-LLM oracle, fully
deterministic on airline (DB-state and communicate checks) and predominantly deterministic on retail
(a deterministic DB check gates an LLM-scored natural-language assertion; Appendix~\ref{app:oracle});
for replication we add SimulatorArena \citep{simulatorarena2025} math tutoring
with human satisfaction ratings and human-graded correctness. Per-substrate statistics in Appendix~\ref{app:datastats}. 

\paragraph{Agents.} We evaluate two complementary sets. The \emph{cross-provider ladder}
(\S\ref{sec:crossfamily}, \S\ref{sec:crossprovider}) spans 14 models from six providers (Anthropic, Meta,
Mistral, Qwen, DeepSeek, OpenAI) at two temperatures on the full task pools of 114 retail and 50
airline tasks, yielding 25 scored model-temperature configurations and about 3{,}700 transcripts in
total (Appendix~\ref{app:datastats}). This wide capability span makes ranking validity, and its near-equal
limit, measurable. The
\emph{controlled-degradation set} (\S\ref{sec:freebaseline}, Appendix~\ref{app:degradation}) is a
positive control: 12 Sonnet-4.5 configurations of \emph{a-priori}-known good, medium, or degraded
quality, crossed with 6 strata and 2 domains to give 144 cells and 720 transcripts, degraded only
through inference-time hyperparameters such as temperature, token or step limits, and error tolerance,
with no prompt or code edits. It supplies
the broken-vs-working axis the all-strong grid cannot, so a signal that fails to rank the degraded
configs last exhibits a measurable validity failure \citep{ribeiro2020checklist,adebayo2018sanity}.

\paragraph{Judges and humans.} We score each transcript with two deliberately disjoint satisfaction
rubrics (prompts in Appendix~\ref{app:prompts}). The \emph{gate} is policy-aware: an
operations-supervisor rubric that reads the full transcript (tool calls and goal included) and scores
service quality with policy adherence and task resolution first. The \emph{proxy} is
satisfaction-only: a process-blind first-person shopper (tool calls and task stripped) rating how the
conversation felt, standing in for the human rater gates are usually validated against. Disjoint in role,
evidence, and vocabulary, the two cannot agree by a shared-rubric artifact. Four judges apply the gate
rubric to the full grid (Claude Sonnet-4.5 and Opus-4.8, GPT-5.4 and GPT-5.5), with Opus-4.8 the primary judge \citep{claudeopus48card} and GPT-5.5 its out-of-family
counterpart; results are consistent across all four (\S\ref{sec:crossprovider}). A blind 3-person panel
grounds the scores on a stratified 150-transcript $\tau^2$ sample.

\paragraph{Metrics.} We report agent-level Spearman correlation with model-clustered bootstrap CIs
\citep{efron1979bootstrap}; broken-vs-working AUC; a power analysis
(min-detectable $\rho$); out-of-sample recalibration (leave-one-stratum/agent-out); and panel
reliability as Krippendorff's $\alpha$ \citep{krippendorff1980content} against a human--human ceiling
\citep{artstein2008intercoder}.

\section{Results}

\subsection{Satisfaction does not imply task success}
\label{sec:satnotsuccess}

On a stratified 150-transcript $\tau^2$ sample rated by three blind, independent annotators (a
fully-crossed 150$\times$3 design; Krippendorff's $\alpha{=}0.79$, human--human ceiling \rr$=$0.85),
two independently-built LLM judges reproduce human satisfaction (Opus-4.8
\rr$=$0.846, GPT-5.5 \rr$=$0.827). Yet satisfaction carries essentially no
information about task success: across all 150 transcripts the correlation is flat and non-positive
(\rr$=-$0.147; dose-response curve flat, Fig.~\ref{fig:hero}b). Crucially, this is not an artifact of
the word ``satisfaction'': all five subjective dimensions the panel rated (satisfaction, respect,
clarity, perceived helpfulness, and would-return) are likewise decorrelated from success
($|\rr|\leq0.17$; Appendix~\ref{app:subjective}), so the gap reflects subjective approval in general.
Concretely, 57.5\% of the conversations the panel rated \emph{satisfied} ($\geq$5/7)
had failed the task.

\paragraph{Uninformative, not merely weak.}
Interpreted against the base rate, this pattern is the central finding (Table~\ref{tab:baserate}). The 57.5\%
satisfied-but-failed rate is statistically indistinguishable from the stratified sample's own 57.3\%
base failure rate: conditioning on \emph{satisfied} does not lower
failure, so satisfaction is \emph{uninformative} rather than merely weak, and at the transcript level
it does not discriminate success (AUC 0.44). By contrast, the policy-aware gate assesses
whether the task resolved: it roughly halves failure risk (20.0\% of gate-accepted conversations fail
against a 40.2\% base rate on the natural task mix) and discriminates success (AUC 0.73; per-score calibration in Appendix~\ref{app:calibration}). The process-blind
\emph{proxy}, which never sees the tool calls or task, behaves like satisfaction, not the gate (32.7\% vs.\ 40.2\%; AUC 0.49). Separating these
two signals, one uninformative and one informative, is the core contribution of this work. The panel result is
robust to dropping any single annotator (56.4--60.5\% leave-one-annotator-out;
Appendix~\ref{app:ratercensus}).

\begin{table}[t]
\centering\small
\setlength{\tabcolsep}{4pt}
\begin{tabular}{@{}lrrrr@{}}
\toprule
Signal & \shortstack[r]{P(fail\,$\mid$\\satisf.)} & \shortstack[r]{base\\rate} & RR & AUC \\
\midrule
Human satisfaction (panel) & 57.5 & 57.3 & 1.00 & 0.44 \\
Satisfaction proxy (grid)  & 32.7 & 40.2 & 0.81 & 0.49 \\
\textbf{Policy-aware gate} (grid) & \textbf{20.0} & 40.2 & \textbf{0.50} & \textbf{0.73} \\
\bottomrule
\end{tabular}
\caption{Satisfaction matches the base failure rate; only the policy-aware gate improves on it. For each
signal: the failure rate among the conversations it rated \emph{satisfied}, the pool's own
base failure rate, their ratio (relative risk, RR), and transcript-level discrimination (AUC). The
human panel is a stratified sample (base 57.3\%); the grid signals use the natural task mix (base
40.2\%).}
\label{tab:baserate}
\end{table}

\paragraph{Robust across raters and domains.}
The satisfied-but-failed rate is robust across sample scale, raters, and providers: it is concordant
across five rater populations (the human panel plus gate and proxy scorings from two
LLM providers; 47.6--59.5\%; Fig.~\ref{fig:hero}a,
Table~\ref{tab:ratercensus}) and stable across satisfaction thresholds. It also holds within each
domain: satisfied-but-failed is 24.1\% (retail), 62.1\% (airline), and 38.7\% (math
tutoring), and the ranking replicates per domain (\rr$=$0.95 retail, 0.80 airline;
Appendices~\ref{app:perjudge},~\ref{app:rangerestriction}). The process-blind proxy overstates success more than the policy-aware gate,
which observes whether the task resolved.

\begin{figure}[t]
  \centering
  \includegraphics[width=\columnwidth]{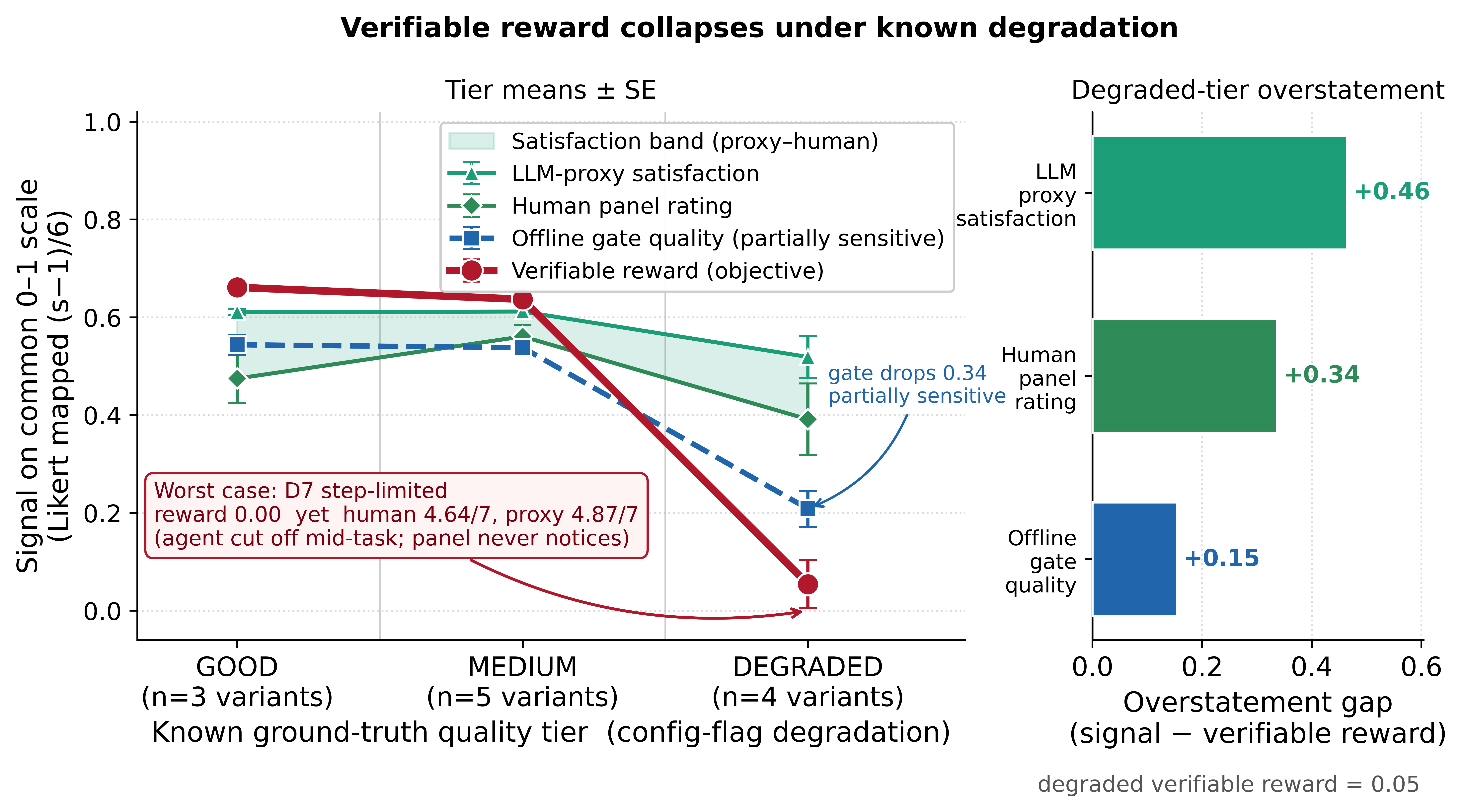}
  \caption{Construct-gap positive control: an agent degraded by configuration flags
  across three known-quality tiers, all signals on a common $[0,1]$ scale.
  As quality drops good$\to$degraded, verifiable reward collapses ($0.66\to0.05$) while the
  satisfaction signals stay nearly flat and the policy-aware gate is only partly sensitive
  ($0.54\to0.21$); the right panel quantifies each signal's overstatement at the degraded tier.}
  \label{fig:pervariant}
\end{figure}

\paragraph{Not a simulator or reward-term artifact.}
The inversion surfaces only where satisfaction is the operative signal: where verifiable reward
already separates agents cleanly, as on the strong-agent six-provider grid, the gate-accept inversion
rate falls to 20\%, the false-accept ceiling a construct-valid gate should meet (292/1{,}460, $p=0.51$). As qualitative cross-substrate corroboration,
it replicates on an independent, human-grounded substrate, SimulatorArena math tutoring, where 38.7\% of
conversations rated $\geq$8/10 by humans were verifiably incorrect ($p=0.013$;
Fig.~\ref{fig:hero}a, right). Swapping our Sonnet-4.5 user-simulator with an independent simulator: GPT-5.4, with agent, tasks, rubrics, and oracle fixed, leaves the inversion intact
(66.7\% gate-satisfied and 69.8\% proxy-satisfied still failing; Appendix~\ref{app:secondsim}). On airline,
whose reward is fully deterministic, the proxy still false-accepts 62.1\% of \emph{satisfied}
conversations (325/523). A degradation positive control demonstrates this directly
(Fig.~\ref{fig:pervariant}): as one agent is degraded across known-quality tiers, reward collapses
while satisfaction stays flat; the starkest case is a transcript truncated mid-task, scoring reward 0.0 yet earning the
highest human rating of any degraded variant (4.6/7; Appendices~\ref{app:degradation},~\ref{app:examples}).

\begin{figure}[t]
  \centering
  \includegraphics[width=\columnwidth]{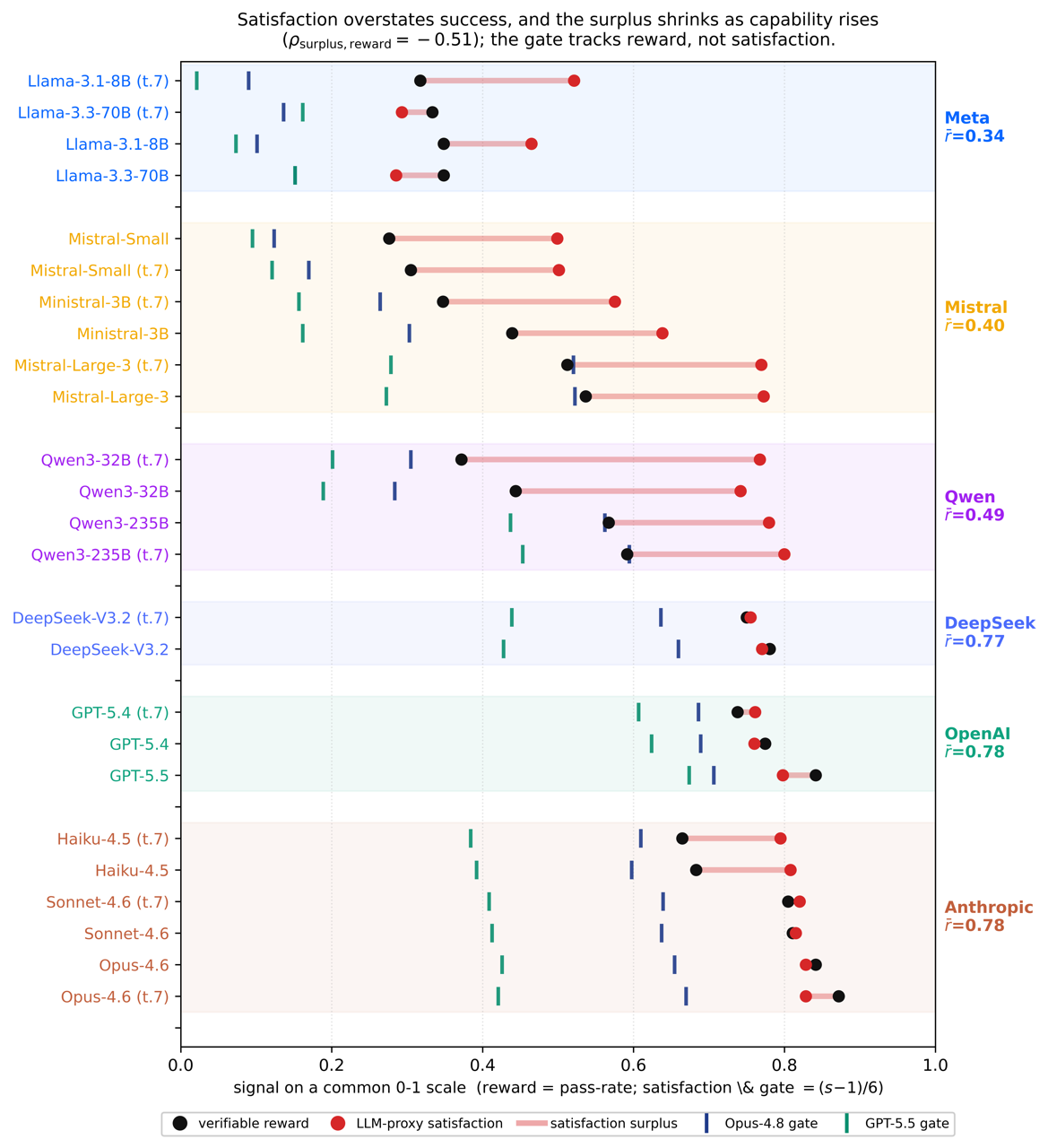}
  \caption{Per-agent verifiable reward, LLM-proxy satisfaction, and the two frontier-judge gates
  (Opus-4.8, GPT-5.5) for all 25 agents. The bar is the \emph{satisfaction surplus}
  (satisfaction${-}$reward). Both gate ticks track reward rather than satisfaction, and every agent
  is rated more satisfying than it succeeds.}
  \label{fig:ladder}
\end{figure}

\subsection{Ranking is robust across providers}
\label{sec:crossfamily}
The construct gap might suggest the gate is invalid, yet its \emph{ranking} is sound. Across the
25-agent, six-provider ladder (Fig.~\ref{fig:ladder}; per-agent values in Appendix~\ref{app:peragent}), the Opus-4.8 gate recovers the
verifiable-reward ordering at \rr$=$0.94, far above the minimum correlation resolvable with $N{=}25$ agents (\rr$=$0.54 at $p{<}0.05$); the
ranking holds across both domains and all four judges and cleanly separates broken from working
agents on the controlled-degradation set (AUC$=$1.00; per-judge and per-domain coefficients in
Appendix~\ref{app:perjudge}). The ordering persists under an independent-provider user-simulator:
re-running the full 25-agent grid with GPT-5.4 (agent and oracle fixed) preserves it at \rr$=$0.93, against 0.94 under Sonnet-4.5, so the ranking
reflects agent quality rather than same-family simulator--agent affinity (Appendix~\ref{app:secondsim}).

This validity is driven by breadth: it lapses among the near-equal strong agents that real release
decisions compare. Such pairs are common (30\% of all pairs and 55\% of top-half pairs differ in
verifiable reward by less than 0.1, our near-equal threshold, near the oracle's resolution given per-agent sampling noise), so we report the
\emph{decision-disagreement rate}: how often the gate promotes the strictly \emph{lower}-reward agent
of a near-equal pair. The primary Opus-4.8 gate does so on 31\% of them versus $<$1\% on wide pairs, and the four other signals behave alike
(GPT-5.4 30\%, GPT-5.5 39\%, human-proxy 30\%, pooled gate 38\%; Appendix~\ref{app:perjudge}), over thirty times the 0.9\% wide-pair rate. This separation is robust
to resampling: the near-equal rate's base-model cluster-bootstrap 95\% CI is $[11.6,50.0]\%$, whose
lower bound alone is ${\sim}12{\times}$ the wide-pair rate. This is a resolution limit at
deployment-relevant separations, not a non-significant correlation. GAUGE thus certifies the gate as
a regression and coarse-quality detector and flags where it cannot adjudicate, a safety-relevant gap
current gate-based evaluation overlooks.

\subsection{Judge robustness and self-preference}
\label{sec:crossprovider}
Because the Anthropic judge and the agents share a provider, ``validity'' could
reflect self-recognition \citep{panickssery2024selfpref}, so we re-score the grid with an
out-of-family judge. Opus-4.8 and GPT-5.5 agree at \rr$=$0.92, all four judges recover the
verifiable-reward ordering (\rr$=$0.84--0.94) and agree with one another (\rr$=$0.90--0.98), and the
ordering is thus not a single-provider artifact, though GPT-5.5 grades $\approx$0.8/7 lower.

The ordering is also not a capability artifact. Run as agents on the same $\tau^2$ pool, both frontier judges are
statistical peers of the strongest agent (Opus-4.8 \passOPUS, GPT-5.5 \passGPT, vs.\ pool-best Opus-4.6 (t.7) at \passTOP; overlapping 95\% CIs); they recover the ordering without
outclassing it, while the weakest judge-as-agent (GPT-5.4, 0.77) is tied-best as a ranker
(\rr$=$0.94). The near-equal limit holds for every judge (top-11 \rr$=$0.25--0.51, all n.s.;
Appendix~\ref{app:frontier}), confirming genuine agent near-equality rather than mis-ranking by a weak judge.

The two-provider design also isolates same-family self-preference: the Opus-4.8 judge inflates Claude
agents over GPT-5.5 more than non-Claude agents, a scale-invariant diff-in-diff of $+0.75$/7
(Sonnet-4.5 shows $+0.67$/7 against GPT-5.4; Appendix~\ref{app:selfpref}), so it reflects
self-preference and leaves the ranking intact. Two reruns confirm the
ranking is stable against judge noise (test--retest ICC$=$0.87, rank stability \rr$=$0.92;
Appendix~\ref{app:judgestability}).

\subsection{Personas stress-test agents}
\label{sec:personas}
The simulator drives each agent against six persona strata spanning cooperativeness, patience, and
assertiveness (axes informed by the OPeRA persona schema \citep{opera2026}; overlays in Appendix~\ref{app:personas}). A single
cooperative user masks failures: relative to the default $\tau^2$ baseline (12 variants, retail;
Fig.~\ref{fig:persona}a), the persona population cuts mean verifiable reward by 42\% ($0.47\to0.27$).

The personas also expose the construct gap (Fig.~\ref{fig:persona}b): satisfaction swings widely by
stratum (proxy $3.5$--$5.4$/7) yet does not track success (the anxious/low-trust and skeptical strata
tie on reward, $0.38$, but differ by $1.2/7$), so a satisfaction-optimized signal tracks who the user
is rather than whether the task succeeded. The benchmark stays stable: every degraded variant floors
at reward $0$ and all working variants stay above (whole-set reward-rank \rr$=$0.83), so the
ranking validity (\S\ref{sec:crossfamily}) reflects agent quality, not the simulator.

\begin{table}[t]
\centering\scriptsize
\setlength{\tabcolsep}{4pt}
\renewcommand{\arraystretch}{1.15}
\begin{tabular}{@{}lccl@{}}
\toprule
Signal & \$/dec. & \rr & Use / cadence \\
\midrule
\textbf{Completion bit} & \textbf{0} & \textbf{0.87} & regressions; every CI run \\
Agent cost / length & 0 & 0.17--0.34 & {--} \\
LLM-judge gate & $+$0.60 & 0.80 & coarse rank; bit insufficient \\
LLM-proxy sat. & $+$0.60 & 0.35 & tone only, \emph{not} success \\
Verifiable audit & rollouts & {--} & calibration; periodic \\
\bottomrule
\end{tabular}
\caption{Release signals by marginal \$/decision, Spearman \rr\ vs.\ the verifiable reward
(controlled-degradation set, $N{=}12$ variants), and recommended use/cadence. The zero-cost
completion bit catches severe regressions as well as the paid judge (\rr\ 0.87 vs.\ 0.80);
agent-cost and proxy \rr\ (0.17--0.35) are not significant.}
\label{tab:deploy}
\end{table}

\subsection{A free signal catches severe regressions}
\label{sec:freebaseline}
A judge-free \emph{completion bit} (conversation completed), parsed at zero cost from
existing rollouts, is a targeted tripwire. On the
controlled-degradation set (broken-vs-working) it scores \rr$=$0.87 against the verifiable reward
versus 0.80 for the full gate (Table~\ref{tab:deploy}), because those failures are truncation-style.
But truncation is the minority failure mode: of 1{,}485 grid failures, 96.5\% terminate normally
(semantic failures) and only 3.5\% truncate, so the bit's recall is 1.0 on truncation but 0.035
overall, and on the six-provider grid it collapses to \rr$=$0.51 versus the gate's 0.94. Semantic
failures need the paid judge, which does carry signal (\rr$=$0.37, AUC 0.72). This division of
labor is the \emph{calibrate-then-trust} recipe: screen routine changes with the free bit for
truncation regressions, and reserve the \$0.60/decision judge for the semantic failures and near-equal
ranking (\S\ref{sec:crossfamily}) it alone resolves.

\subsection{Negative results and additional analyses}
\label{sec:negative}
Recalibration does not transfer, which is why we recommend calibrate-then-trust over a gate
fix. Both natural repair routes fail out-of-sample: a human-anchored persona-reweighting lifts
in-sample validity from \rr$=$0.71 to 0.80, but the per-persona bias term does not hold under leave-one-stratum-out (dropping to \rr$=$0.68); combining all four judges gives no lift over the single best
and is worse under leave-one-agent-out. Both gains are in-sample artifacts
(Appendix~\ref{app:judgestability}): the dependable path is to learn the gate's trusted operating region and
operate within it (\S\ref{sec:freebaseline}), rather than attempting to repair it. Cheap substitutes for the audit fare
no better on the near-equal regime: across 21 candidate signals the four-judge average reproduces the
best single judge's 31\% error exactly (27/87), a common-mode failure (the judges correlate
0.67-0.98 and flip the same pairs; Appendix~\ref{app:cheap}); only \emph{calibrated abstention},
declining pairs whose gate gap is within its sampling noise, reduces it, halving error on the pairs it
ranks (14.8\%).

\section{Discussion and Guidance}
A gate can be human-validated yet still mis-rank on success: ranking and construct validity are
separate properties, and certifying one against human satisfaction does not guarantee the other. Framed positively, satisfaction and task completion are complementary axes (satisfaction validly captures experience), so a release gate should certify both rather than treat one as a proxy for the other. The
cost-reducing remedy is a cadence (Table~\ref{tab:deploy}): run a cheap signal on every config change
and the expensive verifiable audit on a trigger rather than a fixed calendar (a change of judge or
simulator, a shift in the candidate pool, or a near-equal candidate set), to calibrate the
cheap signal rather than replace it.

Three guidelines follow. \textbf{Never gate on
satisfaction alone, especially when optimizing}: once the gate is an optimization target (prompt/policy/model search),
tuning to a gate raises a score decorrelated from
success, so pair any satisfaction signal with a verifiable check.
\textbf{Aggregate to the variant}, never a single (variant$\times$persona) cell. \textbf{When judge
and candidate share a provider, calibrate thresholds against an out-of-family judge}: the $+0.75$/7
self-inflation (\S\ref{sec:crossprovider}) can flip an accept/reject bar even with the ranking intact.

\section*{Limitations}

\paragraph{Human panel size.} The panel comprises 3 annotators rating 150 transcripts in a
fully-crossed design (Krippendorff's $\alpha{=}0.79$, human--human ceiling \rr$=$0.85). While
sufficient for the headline satisfaction--success decorrelation (\rr$=-$0.147, replicated on
SimulatorArena at 38.7\%), a larger panel would tighten per-dimension confidence intervals. The
headline is robust to any single rater: leave-one-annotator-out keeps it within the 56.4--60.5\% band around
57.5\% (Appendix~\ref{app:ratercensus}).

\paragraph{Degradation coverage.} The controlled-degradation set uses budget-style perturbations
(output-token caps, step caps, error-abort; Appendix~\ref{app:degradation}), so its
broken-vs-working result detects truncated agents rather than every failure mode. The full
capability ladder provides complementary coverage of naturalistic failures.

\paragraph{Synthetic substrates.} $\tau^2$-bench user instructions are synthetic, and our absolute
pass-rates are not leaderboard-comparable because the harness runs on Bedrock with a tolerant output
parser (Appendix~\ref{app:oracle}). The parser is lossless by construction, leaving the verifiable
reward's deterministic DB-state and action checks unchanged, but generalization to live customer
interactions remains untested.

\section*{Ethics Statement}
Auditing customer-agent release gates supports safer deployment: satisfied-but-failed inversions are
silent failures that harm customers without surfacing in satisfaction-based monitoring. The
human study rated only de-identified,
synthetic-task transcripts (no personal data) and was therefore IRB-exempt; the three consenting
adult annotators (representative lay raters, independent of the project team) were compensated
above the local minimum wage and rated blind to the AI scores, variant, and outcome.

\bibliography{anthology, references}

@inproceedings{bodhwani2025calibrated,
  title     = {A Calibrated Reflection Approach for Enhancing Confidence Estimation in {LLMs}},
  author    = {Bodhwani, Umesh and Ling, Yuan and Dong, Shujing and Feng, Yarong and Li, Hongfei and Goyal, Ayush},
  booktitle = {Proceedings of the 5th Workshop on Trustworthy NLP (TrustNLP 2025)},
  pages     = {399--411},
  month     = may,
  year      = {2025},
  address   = {Albuquerque, New Mexico},
  publisher = {Association for Computational Linguistics},
  doi       = {10.18653/v1/2025.trustnlp-main.26}
}

@inproceedings{lentex2025,
  title     = {{LentEx}: Generalizable Latent Entity Extraction via Synthetic Data and Instruction-Tuned {LLMs}},
  author    = {Bodhwani, Umesh and Ling, Yuan and Senthilkumar, Cibi Chakravarthy and Dong, Shujing and Feng, Yarong and Li, Hongfei and Goyal, Ayush},
  booktitle = {2025 International Joint Conference on Neural Networks (IJCNN)},
  pages     = {1--8},
  year      = {2025},
  doi       = {10.1109/IJCNN64981.2025.11228382}
}

@inproceedings{selfinstruct2023,
  title     = {{Self-Instruct}: Aligning Language Models with Self-Generated Instructions},
  author    = {Wang, Yizhong and Kordi, Yeganeh and Mishra, Swaroop and Liu, Alisa and Smith, Noah A. and Khashabi, Daniel and Hajishirzi, Hannaneh},
  booktitle = {Proceedings of the 61st Annual Meeting of the Association for Computational Linguistics (Volume 1: Long Papers)},
  pages     = {13484--13508},
  year      = {2023},
  publisher = {Association for Computational Linguistics}
}

@inproceedings{barres2025tau2bench,
  title     = {$\tau^2$-Bench: Evaluating Conversational Agents in a Dual-Control Environment},
  author    = {Barres, Victor and Dong, Honghua and Ray, Soham and Si, Xujie and Narasimhan, Karthik},
  booktitle = {Proceedings of the 43rd International Conference on Machine Learning},
  year      = {2026},
  url       = {https://icml.cc/virtual/2026/oral/71171}
}

@inproceedings{
yao2024taubench,
title={\{\${\textbackslash}tau\$\}-bench: A Benchmark for {\textbackslash}underline\{T\}ool-{\textbackslash}underline\{A\}gent-{\textbackslash}underline\{U\}ser Interaction in Real-World Domains},
author={Shunyu Yao and Noah Shinn and Pedram Razavi and Karthik R Narasimhan},
booktitle={The Thirteenth International Conference on Learning Representations},
year={2025},
url={https://openreview.net/forum?id=roNSXZpUDN}
}

@inproceedings{simulatorarena2025,
  title     = {{SimulatorArena}: Are User Simulators Reliable Proxies for Multi-Turn Evaluation of {AI} Assistants?},
  author    = {Dou, Yao and Galley, Michel and Peng, Baolin and Kedzie, Chris and Cai, Weixin and Ritter, Alan and Quirk, Chris and Xu, Wei and Gao, Jianfeng},
  booktitle = {Proceedings of the 2025 Conference on Empirical Methods in Natural Language Processing},
  pages     = {35212--35290},
  year      = {2025},
  publisher = {Association for Computational Linguistics},
  doi       = {10.18653/v1/2025.emnlp-main.1786}
}

@inproceedings{ecombench2025,
    title = "{EC}om-Bench: Can {LLM} Agent Resolve Real-World {E}-commerce Customer Support Issues?",
    author = "Wang, Haoxin  and
      Peng, Xianhan  and
      Cheng, Huang  and
      Huang, Yizhe  and
      Gong, Ming  and
      Yang, Chenghan  and
      Liu, Yang  and
      Lin, Jiang",
    editor = "Potdar, Saloni  and
      Rojas-Barahona, Lina  and
      Montella, Sebastien",
    booktitle = "Proceedings of the 2025 Conference on Empirical Methods in Natural Language Processing: Industry Track",
    month = nov,
    year = "2025",
    address = "Suzhou (China)",
    publisher = "Association for Computational Linguistics",
    url = "https://aclanthology.org/2025.emnlp-industry.19/",
    doi = "10.18653/v1/2025.emnlp-industry.19",
    pages = "276--284",
    ISBN = "979-8-89176-333-3",
}

@inproceedings{chatbench2025,
  title     = {{ChatBench}: From Static Benchmarks to Human-{AI} Evaluation},
  author    = {Chang, Serina and Anderson, Ashton and Hofman, Jake M.},
  booktitle = {Proceedings of the 63rd Annual Meeting of the Association for Computational Linguistics (Volume 1: Long Papers)},
  pages     = {26009--26038},
  year      = {2025},
  publisher = {Association for Computational Linguistics},
  doi       = {10.18653/v1/2025.acl-long.1262}
}

@inproceedings{opera2026,
  title={{OPeRA}: A Dataset of Observation, Persona, Rationale, and Action for Evaluating {LLMs} on Human Online Shopping Behavior Simulation},
  author={Wang, Ziyi and Lu, Yuxuan and Li, Wenbo and Amini, Amirali and Sun, Bo and Bart, Yakov and Lyu, Weimin and Gesi, Jiri and Wang, Tian and Huang, Jing and Su, Yu and Ehsan, Upol and Alikhani, Malihe and Li, Toby Jia-Jun and Chilton, Lydia and Wang, Dakuo},
  booktitle={Proceedings of the 64th Annual Meeting of the Association for Computational Linguistics (Volume 1: Long Papers)},
  pages={43942--43960},
  year={2026},
  publisher = {Association for Computational Linguistics},
  doi={10.18653/v1/2026.acl-long.2033}
}

@inproceedings{lostinsim2026,
  title     = {Lost in Simulation: {LLM}-Simulated Users are Unreliable Proxies for Human Users in Agentic Evaluations},
  author    = {Seshadri, Preethi and Cahyawijaya, Samuel and Odumakinde, Ayomide and Singh, Sameer and Goldfarb-Tarrant, Seraphina},
  booktitle = {Proceedings of the 64th Annual Meeting of the Association for Computational Linguistics (Volume 1: Long Papers)},
  pages     = {47423--47439},
  year      = {2026},
  publisher = {Association for Computational Linguistics},
  doi       = {10.18653/v1/2026.acl-long.2192}
}

@inproceedings{mindsim2real2026,
  title     = {Mind the Sim2Real Gap in User Simulation for Agentic Tasks},
  author    = {Zhou, Xuhui and Sun, Weiwei and Ma, Qianou and Xie, Yiqing and Liu, Jiarui and Du, Weihua and Welleck, Sean and Yang, Yiming and Neubig, Graham and Wu, Sherry Tongshuang and Sap, Maarten},
  booktitle = {Conference on Language Modeling (COLM)},
  year      = {2026},
  note      = {arXiv:2603.11245}
}

@inproceedings{jung2025trust,
  title     = {Trust or Escalate: {LLM} Judges with Provable Guarantees for Human Agreement},
  author    = {Jung, Jaehun and Brahman, Faeze and Choi, Yejin},
  booktitle = {The Thirteenth International Conference on Learning Representations},
  year      = {2025},
  url       = {https://openreview.net/forum?id=UHPnqSTBPO}
}

@inproceedings{zheng2023judging,
author = {Zheng, Lianmin and Chiang, Wei-Lin and Sheng, Ying and Zhuang, Siyuan and Wu, Zhanghao and Zhuang, Yonghao and Lin, Zi and Li, Zhuohan and Li, Dacheng and Xing, Eric P. and Zhang, Hao and Gonzalez, Joseph E. and Stoica, Ion},
title = {Judging LLM-as-a-judge with MT-bench and Chatbot Arena},
year = {2023},
publisher = {Curran Associates Inc.},
address = {Red Hook, NY, USA},
booktitle = {Proceedings of the 37th International Conference on Neural Information Processing Systems},
articleno = {2020},
numpages = {29},
location = {New Orleans, LA, USA},
series = {NIPS '23}
}

@inproceedings{liu2023geval,
    title = "{G}-Eval: {NLG} Evaluation using Gpt-4 with Better Human Alignment",
    author = "Liu, Yang  and
      Iter, Dan  and
      Xu, Yichong  and
      Wang, Shuohang  and
      Xu, Ruochen  and
      Zhu, Chenguang",
    editor = "Bouamor, Houda  and
      Pino, Juan  and
      Bali, Kalika",
    booktitle = "Proceedings of the 2023 Conference on Empirical Methods in Natural Language Processing",
    month = dec,
    year = "2023",
    address = "Singapore",
    publisher = "Association for Computational Linguistics",
    url = "https://aclanthology.org/2023.emnlp-main.153/",
    doi = "10.18653/v1/2023.emnlp-main.153",
    pages = "2511--2522"
}

@inproceedings{wang2024fair,
    title = "Large Language Models are not Fair Evaluators",
    author = "Wang, Peiyi  and
      Li, Lei  and
      Chen, Liang  and
      Cai, Zefan  and
      Zhu, Dawei  and
      Lin, Binghuai  and
      Cao, Yunbo  and
      Kong, Lingpeng  and
      Liu, Qi  and
      Liu, Tianyu  and
      Sui, Zhifang",
    editor = "Ku, Lun-Wei  and
      Martins, Andre  and
      Srikumar, Vivek",
    booktitle = "Proceedings of the 62nd Annual Meeting of the Association for Computational Linguistics (Volume 1: Long Papers)",
    month = aug,
    year = "2024",
    address = "Bangkok, Thailand",
    publisher = "Association for Computational Linguistics",
    url = "https://aclanthology.org/2024.acl-long.511/",
    doi = "10.18653/v1/2024.acl-long.511",
    pages = "9440--9450"
}

@inproceedings{panickssery2024selfpref,
author = {Panickssery, Arjun and Bowman, Samuel R. and Feng, Shi},
title = {LLM evaluators recognize and favor their own generations},
year = {2024},
isbn = {9798331314385},
publisher = {Curran Associates Inc.},
address = {Red Hook, NY, USA},
booktitle = {Proceedings of the 38th International Conference on Neural Information Processing Systems},
articleno = {2197},
numpages = {31},
location = {Vancouver, BC, Canada},
series = {NIPS '24}
}

@misc{stureborg2024inconsistent,
  title        = {Large Language Models are Inconsistent and Biased Evaluators},
  author       = {Stureborg, Rickard and Alikaniotis, Dimitris and Suhara, Yoshi},
  year         = {2024},
  eprint       = {2405.01724},
  archivePrefix= {arXiv},
  primaryClass = {cs.CL},
}

@misc{gu2024survey,
  title        = {A Survey on {LLM-as-a-Judge}},
  author       = {Gu, Jiawei and Jiang, Xuhui and Shi, Zhichao and Tan, Hexiang and Zhai, Xuehao and Xu, Chengjin and Li, Wei and Shen, Yinghan and Ma, Shengjie and Liu, Honghao and Wang, Saizhuo and Zhang, Kun and Wang, Yuanzhuo and Gao, Wen and Ni, Lionel and Guo, Jian},
  year         = {2024},
  eprint       = {2411.15594},
  archivePrefix= {arXiv},
}

@inproceedings{
tan2025judgebench,
title={JudgeBench: A Benchmark for Evaluating {LLM}-Based Judges},
author={Sijun Tan and Siyuan Zhuang and Kyle Montgomery and William Yuan Tang and Alejandro Cuadron and Chenguang Wang and Raluca Popa and Ion Stoica},
booktitle={The Thirteenth International Conference on Learning Representations},
year={2025},
url={https://openreview.net/forum?id=G0dksFayVq}
}

@inproceedings{sun2021uss,
author = {Sun, Weiwei and Zhang, Shuo and Balog, Krisztian and Ren, Zhaochun and Ren, Pengjie and Chen, Zhumin and de Rijke, Maarten},
title = {Simulating User Satisfaction for the Evaluation of Task-oriented Dialogue Systems},
year = {2021},
isbn = {9781450380379},
publisher = {Association for Computing Machinery},
address = {New York, NY, USA},
url = {https://doi.org/10.1145/3404835.3463241},
doi = {10.1145/3404835.3463241},
booktitle = {Proceedings of the 44th International ACM SIGIR Conference on Research and Development in Information Retrieval},
pages = {2499–2506},
numpages = {8},
location = {Virtual Event, Canada},
series = {SIGIR '21}
}

@misc{davidson2023usersim,
  title        = {User Simulation with Large Language Models for Evaluating Task-Oriented Dialogue},
  author       = {Davidson, Sam and Romeo, Salvatore and Shu, Raphael and Gung, James and Gupta, Arshit and Mansour, Saab and Zhang, Yi},
  year         = {2023},
  eprint       = {2309.13233},
  archivePrefix= {arXiv},
}

@inproceedings{sekulic2024reliable,
    title = "Reliable {LLM}-based User Simulator for Task-Oriented Dialogue Systems",
    author = "Sekulic, Ivan  and
      Terragni, Silvia  and
      Guimar{\~a}es, Victor  and
      Khau, Nghia  and
      Guedes, Bruna  and
      Filipavicius, Modestas  and
      Manso, Andre Ferreira  and
      Mathis, Roland",
    editor = "Graham, Yvette  and
      Liu, Qun  and
      Lampouras, Gerasimos  and
      Iacobacci, Ignacio  and
      Madden, Sinead  and
      Khalid, Haider  and
      Qureshi, Rameez",
    booktitle = "Proceedings of the 1st Workshop on Simulating Conversational Intelligence in Chat (SCI-CHAT 2024)",
    month = mar,
    year = "2024",
    address = "St. Julians, Malta",
    publisher = "Association for Computational Linguistics",
    url = "https://aclanthology.org/2024.scichat-1.3/",
    doi = "10.18653/v1/2024.scichat-1.3",
    pages = "19--35"
}

@InProceedings{zhuge2024agentjudge,
  title = 	 {Agent-as-a-Judge: Evaluate Agents with Agents},
  author =       {Zhuge, Mingchen and Zhao, Changsheng and Ashley, Dylan R. and Wang, Wenyi and Khizbullin, Dmitrii and Xiong, Yunyang and Liu, Zechun and Chang, Ernie and Krishnamoorthi, Raghuraman and Tian, Yuandong and Shi, Yangyang and Chandra, Vikas and Schmidhuber, J\"{u}rgen},
  booktitle = 	 {Proceedings of the 42nd International Conference on Machine Learning},
  pages = 	 {80569--80611},
  year = 	 {2025},
  editor = 	 {Singh, Aarti and Fazel, Maryam and Hsu, Daniel and Lacoste-Julien, Simon and Berkenkamp, Felix and Maharaj, Tegan and Wagstaff, Kiri and Zhu, Jerry},
  volume = 	 {267},
  series = 	 {Proceedings of Machine Learning Research},
  month = 	 {13--19 Jul},
  publisher =    {PMLR},
  url = 	 {https://proceedings.mlr.press/v267/zhuge25a.html}
}

@inproceedings{zhou2024webarena,
  title     = {{WebArena}: A Realistic Web Environment for Building Autonomous Agents},
  author    = {Zhou, Shuyan and Xu, Frank F. and Zhu, Hao and Zhou, Xuhui and Lo, Robert and Sridhar, Abishek and Cheng, Xianyi and Ou, Tianyue and Bisk, Yonatan and Fried, Daniel and Alon, Uri and Neubig, Graham},
  booktitle = {International Conference on Learning Representations (ICLR)},
  year      = {2024},
  note      = {arXiv:2307.13854}
}

@inproceedings{liu2024agentbench,
  title     = {{AgentBench}: Evaluating {LLMs} as Agents},
  author    = {Liu, Xiao and Yu, Hao and Zhang, Hanchen and Xu, Yifan and Lei, Xuanyu and Lai, Hanyu and Gu, Yu and Ding, Hangliang and Men, Kaiwen and Yang, Kejuan and Zhang, Shudan and Deng, Xiang and Zeng, Aohan and Du, Zhengxiao and Zhang, Chenhui and Shen, Sheng and Zhang, Tianjun and Su, Yu and Sun, Huan and Huang, Minlie and Dong, Yuxiao and Tang, Jie},
  booktitle = {International Conference on Learning Representations (ICLR)},
  year      = {2024}
}

@inproceedings{
qin2024toolllm,
title={Tool{LLM}: Facilitating Large Language Models to Master 16000+ Real-world {API}s},
author={Yujia Qin and Shihao Liang and Yining Ye and Kunlun Zhu and Lan Yan and Yaxi Lu and Yankai Lin and Xin Cong and Xiangru Tang and Bill Qian and Sihan Zhao and Lauren Hong and Runchu Tian and Ruobing Xie and Jie Zhou and Mark Gerstein and dahai li and Zhiyuan Liu and Maosong Sun},
booktitle={The Twelfth International Conference on Learning Representations},
year={2024},
url={https://openreview.net/forum?id=dHng2O0Jjr}
}

@inproceedings{lambert2024rewardbench,
    title = "{R}eward{B}ench: Evaluating Reward Models for Language Modeling",
    author = "Lambert, Nathan  and
      Pyatkin, Valentina  and
      Morrison, Jacob  and
      Miranda, LJ  and
      Lin, Bill Yuchen  and
      Chandu, Khyathi  and
      Dziri, Nouha  and
      Kumar, Sachin  and
      Zick, Tom  and
      Choi, Yejin  and
      Smith, Noah A.  and
      Hajishirzi, Hannaneh",
    editor = "Chiruzzo, Luis  and
      Ritter, Alan  and
      Wang, Lu",
    booktitle = "Findings of the Association for Computational Linguistics: NAACL 2025",
    month = apr,
    year = "2025",
    address = "Albuquerque, New Mexico",
    publisher = "Association for Computational Linguistics",
    url = "https://aclanthology.org/2025.findings-naacl.96/",
    doi = "10.18653/v1/2025.findings-naacl.96",
    pages = "1755--1797",
    ISBN = "979-8-89176-195-7"
}

@article{cronbach1955construct,
  title={Construct validity in psychological tests.},
  author={Lee Joseph Cronbach and Paul E. Meehl},
  journal={Psychological bulletin},
  year={1955},
  volume={52 4},
  pages={
          281-302
        },
  url={https://api.semanticscholar.org/CorpusID:5312179}
}

@inproceedings{jacobs2021measurement,
  title     = {Measurement and Fairness},
  author    = {Jacobs, Abigail Z. and Wallach, Hanna},
  booktitle = {Proceedings of the 2021 ACM Conference on Fairness, Accountability, and Transparency (FAccT)},
  year      = {2021},
  note      = {DOI:10.1145/3442188.3445901. arXiv:1912.05511}
}

@inproceedings{bowman2021benchmarking,
    title = "What Will it Take to Fix Benchmarking in Natural Language Understanding?",
    author = "Bowman, Samuel R.  and
      Dahl, George",
    editor = "Toutanova, Kristina  and
      Rumshisky, Anna  and
      Zettlemoyer, Luke  and
      Hakkani-Tur, Dilek  and
      Beltagy, Iz  and
      Bethard, Steven  and
      Cotterell, Ryan  and
      Chakraborty, Tanmoy  and
      Zhou, Yichao",
    booktitle = "Proceedings of the 2021 Conference of the North American Chapter of the Association for Computational Linguistics: Human Language Technologies",
    month = jun,
    year = "2021",
    address = "Online",
    publisher = "Association for Computational Linguistics",
    url = "https://aclanthology.org/2021.naacl-main.385/",
    doi = "10.18653/v1/2021.naacl-main.385",
    pages = "4843--4855"
}

@article{strathern1997improving,
  title={‘Improving ratings’: audit in the British University system},
  author={Marilyn Strathern},
  journal={European Review},
  year={1997},
  volume={5},
  pages={305 - 321},
  url={https://api.semanticscholar.org/CorpusID:145644958}
}

@misc{amodei2016concrete,
  title        = {Concrete Problems in {AI} Safety},
  author       = {Amodei, Dario and Olah, Chris and Steinhardt, Jacob and Christiano, Paul and Schulman, John and Man\'{e}, Dan},
  year         = {2016},
  eprint       = {1606.06565},
  archivePrefix= {arXiv},
}

@inproceedings{skalse2022reward,
author = {Skalse, Joar and Howe, Nikolaus H. R. and Krasheninnikov, Dmitrii and Krueger, David},
title = {Defining and characterizing reward hacking},
year = {2022},
isbn = {9781713871088},
publisher = {Curran Associates Inc.},
address = {Red Hook, NY, USA},
booktitle = {Proceedings of the 36th International Conference on Neural Information Processing Systems},
articleno = {687},
numpages = {12},
location = {New Orleans, LA, USA},
series = {NIPS '22}
}

@article{shrout1979intraclass,
  title     = {Intraclass Correlations: Uses in Assessing Rater Reliability},
  author    = {Shrout, Patrick E. and Fleiss, Joseph L.},
  journal   = {Psychological Bulletin},
  volume    = {86},
  number    = {2},
  pages     = {420--428},
  year      = {1979},
  note      = {DOI:10.1037/0033-2909.86.2.420}
}

@book{krippendorff1980content,
  title     = {Content Analysis: An Introduction to Its Methodology},
  author    = {Krippendorff, Klaus},
  publisher = {Sage},
  year      = {1980}
}

@article{artstein2008intercoder,
    title = "Survey Article: Inter-Coder Agreement for Computational Linguistics",
    author = "Artstein, Ron  and
      Poesio, Massimo",
    journal = "Computational Linguistics",
    volume = "34",
    number = "4",
    year = "2008",
    url = "https://aclanthology.org/J08-4004/",
    doi = "10.1162/coli.07-034-R2",
    pages = "555--596"
}

@article{efron1979bootstrap,
  title     = {Bootstrap Methods: Another Look at the Jackknife},
  author    = {Efron, Bradley},
  journal   = {The Annals of Statistics},
  volume    = {7},
  number    = {1},
  pages     = {1--26},
  year      = {1979},
  note      = {DOI:10.1214/aos/1176344552}
}

@misc{deepseekv3,
  title        = {{DeepSeek-V3} Technical Report},
  author       = {{DeepSeek-AI}},
  year         = {2024},
  eprint       = {2412.19437},
  archivePrefix= {arXiv},
}

@misc{mistral3,
  title        = {Introducing {Mistral 3}},
  author       = {{Mistral AI}},
  year         = {2025},
  month        = dec,
  howpublished = {\url{https://mistral.ai/news/mistral-3/}},
  note         = {Official release announcement for Mistral Large 3 and the Ministral 3 family}
}

@misc{openai2026gpt54,
  title        = {Introducing {GPT-5.4}},
  author       = {{OpenAI}},
  year         = {2026},
  howpublished = {\url{https://openai.com/index/introducing-gpt-5-4/}},
  note         = {Official release announcement}
}

@misc{openai2026gpt55,
  title        = {Introducing {GPT-5.5}},
  author       = {{OpenAI}},
  year         = {2026},
  howpublished = {\url{https://openai.com/index/introducing-gpt-5-5/}},
  note         = {Official release announcement}
}

@misc{claudeopus48card,
  title        = {{Claude} {Opus} 4.8 System Card},
  author       = {{Anthropic}},
  year         = {2026},
  howpublished = {\url{https://www.anthropic.com/claude-opus-4-8-system-card}},
  note         = {Official system card}
}

@misc{llama33modelcard,
  title        = {{Llama 3.3 70B Instruct} Model Card},
  author       = {{Meta}},
  year         = {2024},
  howpublished = {\url{https://huggingface.co/meta-llama/Llama-3.3-70B-Instruct}},
  note         = {Official model card}
}

@misc{deepseekv32modelcard,
  title        = {{DeepSeek-V3.2} Model Card},
  author       = {{DeepSeek-AI}},
  year         = {2025},
  howpublished = {\url{https://huggingface.co/deepseek-ai/DeepSeek-V3.2}},
  note         = {Official model card}
}

@misc{claudeopus46card,
  title        = {{Claude} {Opus} 4.6 System Card},
  author       = {{Anthropic}},
  year         = {2026},
  howpublished = {\url{https://www.anthropic.com/claude-opus-4-6-system-card}},
  note         = {Official system card}
}

@misc{claudesonnet46card,
  title        = {{Claude} {Sonnet} 4.6 System Card},
  author       = {{Anthropic}},
  year         = {2026},
  howpublished = {\url{https://www.anthropic.com/claude-sonnet-4-6-system-card}},
  note         = {Official system card}
}

@misc{claudehaiku45card,
  title        = {{Claude} {Haiku} 4.5 System Card},
  author       = {{Anthropic}},
  year         = {2025},
  howpublished = {\url{https://www.anthropic.com/claude-haiku-4-5-system-card}},
  note         = {Official system card}
}

@inproceedings{ribeiro2020checklist,
    title = "Beyond Accuracy: Behavioral Testing of {NLP} Models with {C}heck{L}ist",
    author = "Ribeiro, Marco Tulio  and
      Wu, Tongshuang  and
      Guestrin, Carlos  and
      Singh, Sameer",
    editor = "Jurafsky, Dan  and
      Chai, Joyce  and
      Schluter, Natalie  and
      Tetreault, Joel",
    booktitle = "Proceedings of the 58th Annual Meeting of the Association for Computational Linguistics",
    month = jul,
    year = "2020",
    publisher = "Association for Computational Linguistics",
    url = "https://aclanthology.org/2020.acl-main.442/",
    doi = "10.18653/v1/2020.acl-main.442",
    pages = "4902--4912"
}

@inproceedings{adebayo2018sanity,
author = {Adebayo, Julius and Gilmer, Justin and Muelly, Michael and Goodfellow, Ian and Hardt, Moritz and Kim, Been},
title = {Sanity checks for saliency maps},
year = {2018},
publisher = {Curran Associates Inc.},
address = {Red Hook, NY, USA},
booktitle = {Proceedings of the 32nd International Conference on Neural Information Processing Systems},
pages = {9525–9536},
numpages = {12},
location = {Montr{\'e}al, Canada},
series = {NIPS'18}
}

@misc{grattafiori2024llama3herdmodels,
      title={The {Llama} 3 Herd of Models},
      author={Aaron Grattafiori and Abhimanyu Dubey and Abhinav Jauhri and Abhinav Pandey and Abhishek Kadian and Ahmad Al-Dahle and Aiesha Letman and Akhil Mathur and Alan Schelten and Alex Vaughan and Amy Yang and Angela Fan and Anirudh Goyal and Anthony Hartshorn and Aobo Yang and Archi Mitra and Archie Sravankumar and Artem Korenev and Arthur Hinsvark and Arun Rao and Aston Zhang and Aurelien Rodriguez and Austen Gregerson and Ava Spataru and Baptiste Roziere and Bethany Biron and Binh Tang and Bobbie Chern and Charlotte Caucheteux and Chaya Nayak and Chloe Bi and Chris Marra and Chris McConnell and Christian Keller and Christophe Touret and Chunyang Wu and Corinne Wong and Cristian Canton Ferrer and Cyrus Nikolaidis and Damien Allonsius and Daniel Song and Danielle Pintz and Danny Livshits and Danny Wyatt and David Esiobu and Dhruv Choudhary and Dhruv Mahajan and Diego Garcia-Olano and Diego Perino and Dieuwke Hupkes and Egor Lakomkin and Ehab AlBadawy and Elina Lobanova and Emily Dinan and Eric Michael Smith and Filip Radenovic and Francisco Guzmán and Frank Zhang and Gabriel Synnaeve and Gabrielle Lee and Georgia Lewis Anderson and Govind Thattai and Graeme Nail and Gregoire Mialon and Guan Pang and Guillem Cucurell and Hailey Nguyen and Hannah Korevaar and Hu Xu and Hugo Touvron and Iliyan Zarov and Imanol Arrieta Ibarra and Isabel Kloumann and Ishan Misra and Ivan Evtimov and Jack Zhang and Jade Copet and Jaewon Lee and Jan Geffert and Jana Vranes and Jason Park and Jay Mahadeokar and Jeet Shah and Jelmer van der Linde and Jennifer Billock and Jenny Hong and Jenya Lee and Jeremy Fu and Jianfeng Chi and Jianyu Huang and Jiawen Liu and Jie Wang and Jiecao Yu and Joanna Bitton and Joe Spisak and Jongsoo Park and Joseph Rocca and Joshua Johnstun and Joshua Saxe and Junteng Jia and Kalyan Vasuden Alwala and Karthik Prasad and Kartikeya Upasani and Kate Plawiak and Ke Li and Kenneth Heafield and Kevin Stone and Khalid El-Arini and Krithika Iyer and Kshitiz Malik and Kuenley Chiu and Kunal Bhalla and Kushal Lakhotia and Lauren Rantala-Yeary and Laurens van der Maaten and Lawrence Chen and Liang Tan and Liz Jenkins and Louis Martin and Lovish Madaan and Lubo Malo and Lukas Blecher and Lukas Landzaat and Luke de Oliveira and Madeline Muzzi and Mahesh Pasupuleti and Mannat Singh and Manohar Paluri and Marcin Kardas and Maria Tsimpoukelli and Mathew Oldham and Mathieu Rita and Maya Pavlova and Melanie Kambadur and Mike Lewis and Min Si and Mitesh Kumar Singh and Mona Hassan and Naman Goyal and Narjes Torabi and Nikolay Bashlykov and Nikolay Bogoychev and Niladri Chatterji and Ning Zhang and Olivier Duchenne and Onur Çelebi and Patrick Alrassy and Pengchuan Zhang and Pengwei Li and Petar Vasic and Peter Weng and Prajjwal Bhargava and Pratik Dubal and Praveen Krishnan and Punit Singh Koura and Puxin Xu and Qing He and Qingxiao Dong and Ragavan Srinivasan and Raj Ganapathy and Ramon Calderer and Ricardo Silveira Cabral and Robert Stojnic and Roberta Raileanu and Rohan Maheswari and Rohit Girdhar and Rohit Patel and Romain Sauvestre and Ronnie Polidoro and Roshan Sumbaly and Ross Taylor and Ruan Silva and Rui Hou and Rui Wang and Saghar Hosseini and Sahana Chennabasappa and Sanjay Singh and Sean Bell and Seohyun Sonia Kim and Sergey Edunov and Shaoliang Nie and Sharan Narang and Sharath Raparthy and Sheng Shen and Shengye Wan and Shruti Bhosale and Shun Zhang and Simon Vandenhende and Soumya Batra and Spencer Whitman and Sten Sootla and Stephane Collot and Suchin Gururangan and Sydney Borodinsky and Tamar Herman and Tara Fowler and Tarek Sheasha and Thomas Georgiou and Thomas Scialom and Tobias Speckbacher and Todor Mihaylov and Tong Xiao and Ujjwal Karn and Vedanuj Goswami and Vibhor Gupta and Vignesh Ramanathan and Viktor Kerkez and Vincent Gonguet and Virginie Do and Vish Vogeti and Vítor Albiero and Vladan Petrovic and Weiwei Chu and Wenhan Xiong and Wenyin Fu and Whitney Meers and Xavier Martinet and Xiaodong Wang and Xiaofang Wang and Xiaoqing Ellen Tan and Xide Xia and Xinfeng Xie and Xuchao Jia and Xuewei Wang and Yaelle Goldschlag and Yashesh Gaur and Yasmine Babaei and Yi Wen and Yiwen Song and Yuchen Zhang and Yue Li and Yuning Mao and Zacharie Delpierre Coudert and Zheng Yan and Zhengxing Chen and Zoe Papakipos and Aaditya Singh and Aayushi Srivastava and Abha Jain and Adam Kelsey and Adam Shajnfeld and Adithya Gangidi and Adolfo Victoria and Ahuva Goldstand and Ajay Menon and Ajay Sharma and Alex Boesenberg and Alexei Baevski and Allie Feinstein and Amanda Kallet and Amit Sangani and Amos Teo and Anam Yunus and Andrei Lupu and Andres Alvarado and Andrew Caples and Andrew Gu and Andrew Ho and Andrew Poulton and Andrew Ryan and Ankit Ramchandani and Annie Dong and Annie Franco and Anuj Goyal and Aparajita Saraf and Arkabandhu Chowdhury and Ashley Gabriel and Ashwin Bharambe and Assaf Eisenman and Azadeh Yazdan and Beau James and Ben Maurer and Benjamin Leonhardi and Bernie Huang and Beth Loyd and Beto De Paola and Bhargavi Paranjape and Bing Liu and Bo Wu and Boyu Ni and Braden Hancock and Bram Wasti and Brandon Spence and Brani Stojkovic and Brian Gamido and Britt Montalvo and Carl Parker and Carly Burton and Catalina Mejia and Ce Liu and Changhan Wang and Changkyu Kim and Chao Zhou and Chester Hu and Ching-Hsiang Chu and Chris Cai and Chris Tindal and Christoph Feichtenhofer and Cynthia Gao and Damon Civin and Dana Beaty and Daniel Kreymer and Daniel Li and David Adkins and David Xu and Davide Testuggine and Delia David and Devi Parikh and Diana Liskovich and Didem Foss and Dingkang Wang and Duc Le and Dustin Holland and Edward Dowling and Eissa Jamil and Elaine Montgomery and Eleonora Presani and Emily Hahn and Emily Wood and Eric-Tuan Le and Erik Brinkman and Esteban Arcaute and Evan Dunbar and Evan Smothers and Fei Sun and Felix Kreuk and Feng Tian and Filippos Kokkinos and Firat Ozgenel and Francesco Caggioni and Frank Kanayet and Frank Seide and Gabriela Medina Florez and Gabriella Schwarz and Gada Badeer and Georgia Swee and Gil Halpern and Grant Herman and Grigory Sizov and Guangyi and Zhang and Guna Lakshminarayanan and Hakan Inan and Hamid Shojanazeri and Han Zou and Hannah Wang and Hanwen Zha and Haroun Habeeb and Harrison Rudolph and Helen Suk and Henry Aspegren and Hunter Goldman and Hongyuan Zhan and Ibrahim Damlaj and Igor Molybog and Igor Tufanov and Ilias Leontiadis and Irina-Elena Veliche and Itai Gat and Jake Weissman and James Geboski and James Kohli and Janice Lam and Japhet Asher and Jean-Baptiste Gaya and Jeff Marcus and Jeff Tang and Jennifer Chan and Jenny Zhen and Jeremy Reizenstein and Jeremy Teboul and Jessica Zhong and Jian Jin and Jingyi Yang and Joe Cummings and Jon Carvill and Jon Shepard and Jonathan McPhie and Jonathan Torres and Josh Ginsburg and Junjie Wang and Kai Wu and Kam Hou U and Karan Saxena and Kartikay Khandelwal and Katayoun Zand and Kathy Matosich and Kaushik Veeraraghavan and Kelly Michelena and Keqian Li and Kiran Jagadeesh and Kun Huang and Kunal Chawla and Kyle Huang and Lailin Chen and Lakshya Garg and Lavender A and Leandro Silva and Lee Bell and Lei Zhang and Liangpeng Guo and Licheng Yu and Liron Moshkovich and Luca Wehrstedt and Madian Khabsa and Manav Avalani and Manish Bhatt and Martynas Mankus and Matan Hasson and Matthew Lennie and Matthias Reso and Maxim Groshev and Maxim Naumov and Maya Lathi and Meghan Keneally and Miao Liu and Michael L. Seltzer and Michal Valko and Michelle Restrepo and Mihir Patel and Mik Vyatskov and Mikayel Samvelyan and Mike Clark and Mike Macey and Mike Wang and Miquel Jubert Hermoso and Mo Metanat and Mohammad Rastegari and Munish Bansal and Nandhini Santhanam and Natascha Parks and Natasha White and Navyata Bawa and Nayan Singhal and Nick Egebo and Nicolas Usunier and Nikhil Mehta and Nikolay Pavlovich Laptev and Ning Dong and Norman Cheng and Oleg Chernoguz and Olivia Hart and Omkar Salpekar and Ozlem Kalinli and Parkin Kent and Parth Parekh and Paul Saab and Pavan Balaji and Pedro Rittner and Philip Bontrager and Pierre Roux and Piotr Dollar and Polina Zvyagina and Prashant Ratanchandani and Pritish Yuvraj and Qian Liang and Rachad Alao and Rachel Rodriguez and Rafi Ayub and Raghotham Murthy and Raghu Nayani and Rahul Mitra and Rangaprabhu Parthasarathy and Raymond Li and Rebekkah Hogan and Robin Battey and Rocky Wang and Russ Howes and Ruty Rinott and Sachin Mehta and Sachin Siby and Sai Jayesh Bondu and Samyak Datta and Sara Chugh and Sara Hunt and Sargun Dhillon and Sasha Sidorov and Satadru Pan and Saurabh Mahajan and Saurabh Verma and Seiji Yamamoto and Sharadh Ramaswamy and Shaun Lindsay and Shaun Lindsay and Sheng Feng and Shenghao Lin and Shengxin Cindy Zha and Shishir Patil and Shiva Shankar and Shuqiang Zhang and Shuqiang Zhang and Sinong Wang and Sneha Agarwal and Soji Sajuyigbe and Soumith Chintala and Stephanie Max and Stephen Chen and Steve Kehoe and Steve Satterfield and Sudarshan Govindaprasad and Sumit Gupta and Summer Deng and Sungmin Cho and Sunny Virk and Suraj Subramanian and Sy Choudhury and Sydney Goldman and Tal Remez and Tamar Glaser and Tamara Best and Thilo Koehler and Thomas Robinson and Tianhe Li and Tianjun Zhang and Tim Matthews and Timothy Chou and Tzook Shaked and Varun Vontimitta and Victoria Ajayi and Victoria Montanez and Vijai Mohan and Vinay Satish Kumar and Vishal Mangla and Vlad Ionescu and Vlad Poenaru and Vlad Tiberiu Mihailescu and Vladimir Ivanov and Wei Li and Wenchen Wang and Wenwen Jiang and Wes Bouaziz and Will Constable and Xiaocheng Tang and Xiaojian Wu and Xiaolan Wang and Xilun Wu and Xinbo Gao and Yaniv Kleinman and Yanjun Chen and Ye Hu and Ye Jia and Ye Qi and Yenda Li and Yilin Zhang and Ying Zhang and Yossi Adi and Youngjin Nam and Yu and Wang and Yu Zhao and Yuchen Hao and Yundi Qian and Yunlu Li and Yuzi He and Zach Rait and Zachary DeVito and Zef Rosnbrick and Zhaoduo Wen and Zhenyu Yang and Zhiwei Zhao and Zhiyu Ma},
      year={2024},
      eprint={2407.21783},
      archivePrefix={arXiv},
      primaryClass={cs.AI},
      url={https://arxiv.org/abs/2407.21783}, 
}

@misc{yang2025qwen3technicalreport,
      title={Qwen3 Technical Report}, 
      author={An Yang and Anfeng Li and Baosong Yang and Beichen Zhang and Binyuan Hui and Bo Zheng and Bowen Yu and Chang Gao and Chengen Huang and Chenxu Lv and Chujie Zheng and Dayiheng Liu and Fan Zhou and Fei Huang and Feng Hu and Hao Ge and Haoran Wei and Huan Lin and Jialong Tang and Jian Yang and Jianhong Tu and Jianwei Zhang and Jianxin Yang and Jiaxi Yang and Jing Zhou and Jingren Zhou and Junyang Lin and Kai Dang and Keqin Bao and Kexin Yang and Le Yu and Lianghao Deng and Mei Li and Mingfeng Xue and Mingze Li and Pei Zhang and Peng Wang and Qin Zhu and Rui Men and Ruize Gao and Shixuan Liu and Shuang Luo and Tianhao Li and Tianyi Tang and Wenbiao Yin and Xingzhang Ren and Xinyu Wang and Xinyu Zhang and Xuancheng Ren and Yang Fan and Yang Su and Yichang Zhang and Yinger Zhang and Yu Wan and Yuqiong Liu and Zekun Wang and Zeyu Cui and Zhenru Zhang and Zhipeng Zhou and Zihan Qiu},
      year={2025},
      eprint={2505.09388},
      archivePrefix={arXiv},
      primaryClass={cs.CL},
      url={https://arxiv.org/abs/2505.09388}, 
}

\appendix

\section{Satisfied-but-Failed Rates by Rater Population}
\label{app:ratercensus}
Table~\ref{tab:ratercensus} reports the satisfied-but-failed rate behind
Figure~\ref{fig:hero}a and the \S\ref{sec:satnotsuccess} claim: for each independent rater
population, the share of conversations it rated \emph{satisfied} that the non-LLM oracle scored as a
task failure, with an exact one-sided binomial test against a lenient $20\%$ ceiling, the false-accept
rate a construct-valid \emph{gate} should not exceed; for the satisfaction
signals the primary comparator is the base failure rate (Table~\ref{tab:baserate}), which they sit at.
Every population's satisfied-but-failed rate is far above $20\%$: the human panel and all four LLM
scorings on $\tau^2$ reject the ceiling at $p<10^{-6}$, and the independent SimulatorArena human
ratings at $p{=}0.013$.

\begin{table}[t]
\centering\small
\setlength{\tabcolsep}{4pt}
\renewcommand{\arraystretch}{1.05}
\begin{tabular}{@{}lccr@{}}
\toprule
Rater population & Failed/Sat. & Rate\,\% & $p$ \\
\midrule
\multicolumn{4}{@{}l}{\emph{$\tau^2$-bench customer service}}\\
Human panel (3 raters) & 23/40 & 57.5 & $2{\times}10^{-7}$ \\
Opus-4.8 gate & 30/63 & 47.6 & $8{\times}10^{-7}$ \\
Opus-4.8 proxy & 50/84 & 59.5 & $3{\times}10^{-15}$ \\
GPT-5.5 gate & 29/57 & 50.9 & $2{\times}10^{-7}$ \\
GPT-5.5 proxy & 75/126 & 59.5 & $3{\times}10^{-22}$ \\
\midrule
\multicolumn{4}{@{}l}{\emph{SimulatorArena math tutoring}}\\
Human rating ($\geq$8/10) & 12/31 & 38.7 & $0.013$ \\
\bottomrule
\end{tabular}
\caption{Satisfied-but-failed rate per rater population: of conversations a population rated
satisfied (human/Anthropic-scale $\geq$5/7; the GPT scale, which grades $\approx$1 point lower, $\geq$4/7;
SimulatorArena $\geq$8/10), the percentage the verifiable oracle scored as failed. $p$ is a one-sided
exact binomial test against the lenient $20\%$ false-accept ceiling a construct-valid gate would meet.
Every population exceeds it; the gap is not a single-rater, single-provider, or single-substrate artifact.}
\label{tab:ratercensus}
\end{table}

\paragraph{Leave-one-annotator-out.} The panel headline does not depend on any single rater: removing
any one of the three annotators keeps the satisfied-but-failed rate within 56.4--60.5\% (a
4.1-point spread around the reported 57.5\%), and each annotator independently shows the same
inversion (58.1\%, 63.5\%, 59.5\% satisfied-but-failed).

\section{The Construct Gap Is Not Specific to Satisfaction}
\label{app:subjective}
The main text reports \emph{satisfaction} as the primary subjective signal because it is the
construct the release gate's rubric and the LLM-proxy optimize, and the target the LLM-judge
literature validates against. The blind human panel, however, rated \emph{five} subjective
dimensions per transcript (satisfaction, respect/tone, clarity, perceived helpfulness, and
would-return). Table~\ref{tab:subjective} shows the construct gap generalizes beyond
satisfaction: every dimension shows no association with
verifiable task success (all $|\rr|\leq0.17$), and each inverts at a comparable rate (56--64\% of
highly-rated conversations failed). The dimensions are highly collinear (pairwise Spearman
0.61--0.97): they are not five independent tests but roughly one latent ``this interaction felt
good'' factor, measured five ways, that is orthogonal to task success, which is why the lone
nominally significant correlation (would-return, $p{=}0.04$, and \emph{negative}) is a weak effect and, as the only nominal hit among five correlated tests, consistent with chance. Satisfaction is, if anything, the conservative choice: its inversion rate (57.5\%) is among the lowest of the five. None of the other subjective dimensions would close the gap.

\begin{table}[t]
\centering\small
\setlength{\tabcolsep}{3pt}
\begin{tabular}{@{}lrrrrr@{}}
\toprule
Human dimension & mean & \rr$_{\text{succ}}$ & $p$ & \shortstack[r]{high-\\fail\,\%} & $\alpha$ \\
\midrule
\textbf{Satisfaction} & 3.88 & $-0.147$ & 0.07 & 57.5 & 0.79 \\
Respect / tone & 5.26 & $-0.118$ & 0.15 & 63.6 & 0.76 \\
Clarity & 5.02 & $+0.085$ & 0.30 & 56.5 & 0.83 \\
Perceived helpfulness & 4.10 & $-0.129$ & 0.12 & 63.3 & 0.80 \\
Would return & 4.19 & $-0.168$ & 0.04 & 61.7 & 0.81 \\
\bottomrule
\end{tabular}
\caption{Every subjective dimension the human panel rated, not satisfaction alone, is decorrelated
from verifiable success ($n{=}150$). \rr$_{\text{succ}}$ is the Spearman correlation of the
3-annotator consensus with verifiable reward; ``high-fail \%'' is the percentage of
conversations rated $\geq$5/7 on that dimension that failed the task; $\alpha$ is the inter-annotator Krippendorff's $\alpha$ (interval). Satisfaction (the gate's optimized construct,
bold) is the primary case in the main text and the most conservative.}
\label{tab:subjective}
\end{table}

\section{The Controlled-Degradation Set}
\label{app:degradation}

\paragraph{Purpose and design principle.} The headline cross-provider grid (25 agents) establishes
\emph{external} validity (does the gate rank real, deployable models correctly?), but it cannot
cleanly characterize the gate's behavior on a \emph{known-bad} agent, because frontier models are
all competent. The controlled-degradation set supplies that missing \emph{internal} validity: a
single model (Claude Sonnet-4.5) is run under 12 configurations of a-priori-known quality, so any
rating difference is attributable to agent behavior rather than to provider or
prompt-style confounds. This instantiates the controlled-perturbation paradigm reviewed in
\S\ref{sec:related} \citep{ribeiro2020checklist,adebayo2018sanity}: the
known-degradation ``sanity check'' that crippling a system must register on a valid metric, applied
here to a release gate.

\paragraph{Flag-only degradation.} Crucially, every configuration shares the \emph{identical} agent
prompt and tool set; we perturb only inference-time flags: sampling temperature, the
output-token cap (\texttt{max\_tokens}), the conversation step cap (\texttt{max\_steps}), and the
orchestrator error tolerance (\texttt{max\_errors}). No prompt is rewritten and no code is edited.
Each degradation is a misconfiguration a real deployment could plausibly ship: a too-low
output cap truncates tool-call JSON and confirmations mid-emission; a too-low step cap cuts the
trajectory off before the verifiable action (\,$\tau^2$ retail tasks require
authenticate~$\rightarrow$~locate~$\rightarrow$~act~$\rightarrow$~confirm\,);
an aggressive error-abort terminates the episode on the first malformed tool call. This keeps the
intervention transparent and reproducible while breaking verifiable task completion rather
than surface fluency, precisely the regime in which satisfaction can diverge from success.
Because these are \emph{truncation-style} failures, the judge-free completion bit tracks them well
here (\rr$=$0.87; \S\ref{sec:freebaseline}); on the naturalistic six-provider grid, where 96.5\% of
failures instead terminate normally (semantic), the same bit collapses, which is why we scope it as a
truncation-specific tripwire rather than a general regression detector.

\paragraph{Positive control: the ground-truth ordering is known.} By construction the four \texttt{D}
(degraded) configurations must rank below the eight \texttt{A}/\texttt{B} (good/medium) ones. The
non-LLM verifiable reward confirms a large, clean separation (degraded-tier mean reward
\textbf{0.05} versus \textbf{0.65} for good/medium), so the set behaves as a positive control: a
release signal that fails to place the \texttt{D} configs last is exhibiting a measurable validity
failure, not noise. Table~\ref{tab:variants} gives the full specification and every signal's
per-variant mean.

\begin{table}[t]
\centering\small
\setlength{\tabcolsep}{3pt}
\begin{tabular}{@{}llcccc@{}}
\toprule
Config & Perturb. & Rew. & Gate & Prx. & Hum. \\
\midrule
\multicolumn{6}{@{}l}{\emph{Good tier (temperature only; full budgets)}}\\
A1 standard      & $T{=}0$              & 0.67 & 4.08 & 4.62 & 4.15 \\
A2 strict        & $T{=}0$              & 0.63 & 4.20 & 4.63 & 4.15 \\
A3 temp0.2       & $T{=}0.2$            & 0.68 & 4.50 & 4.73 & 3.24 \\
\midrule
\multicolumn{6}{@{}l}{\emph{Medium tier (higher temp; mild caps)}}\\
B3 temp0.7       & $T{=}0.7$            & 0.68 & 4.42 & 4.72 & 4.67 \\
B4 tight-budget  & $T0.7$,tok256        & 0.62 & 4.07 & 4.32 & 4.53 \\
B5 high-var.     & $T{=}1.0$            & 0.65 & 4.17 & 4.87 & 4.25 \\
B6 temp0.5       & $T{=}0.5$            & 0.62 & 4.22 & 4.67 & 4.52 \\
B7 tok384        & $T0.3$,tok384        & 0.62 & 4.27 & 4.78 & 3.85 \\
\midrule
\multicolumn{6}{@{}l}{\emph{Degraded tier (budget/step/error starvation)}}\\
D6 trunc.        & tok96                & 0.20 & 2.27 & 3.68 & 2.94 \\
\textbf{D7 step-starv.} & \textbf{st6}  & \textbf{0.00} & \textbf{2.70} & \textbf{4.87} & \textbf{4.64} \\
D8 compound      & $T1.0$,tok48,err1    & 0.00 & 1.65 & 3.85 & 2.69 \\
D9 tok72/st12    & tok72,st12           & 0.02 & 2.38 & 4.05 & 3.12 \\
\bottomrule
\end{tabular}
\caption{The 12 controlled-degradation configurations (all Claude Sonnet-4.5, flag-only). Columns:
verifiable reward (pass-rate), and mean LLM-judge \textbf{Gate}, LLM-\textbf{Pr}o\textbf{x}y, and
\textbf{Hum}an-panel satisfaction (1--7). \texttt{tok}$=$\texttt{max\_tokens}, \texttt{st}/\texttt{steps}$=$\texttt{max\_steps},
\texttt{err}$=$\texttt{max\_errors}. The degraded tier collapses on verifiable reward
($\approx$0) yet every subjective signal still rates it mid-scale. \textbf{D7} (highlighted) is the
sharpest inversion: reward $0.00$ but human satisfaction $4.64$, the highest of any degraded
config, because a step-capped trajectory is cut off mid-task while reading as helpful turn-by-turn.}
\label{tab:variants}
\end{table}

\subsection{Gate-score calibration}
\label{app:calibration}
\begin{figure}[t]
  \centering
  \includegraphics[width=\columnwidth]{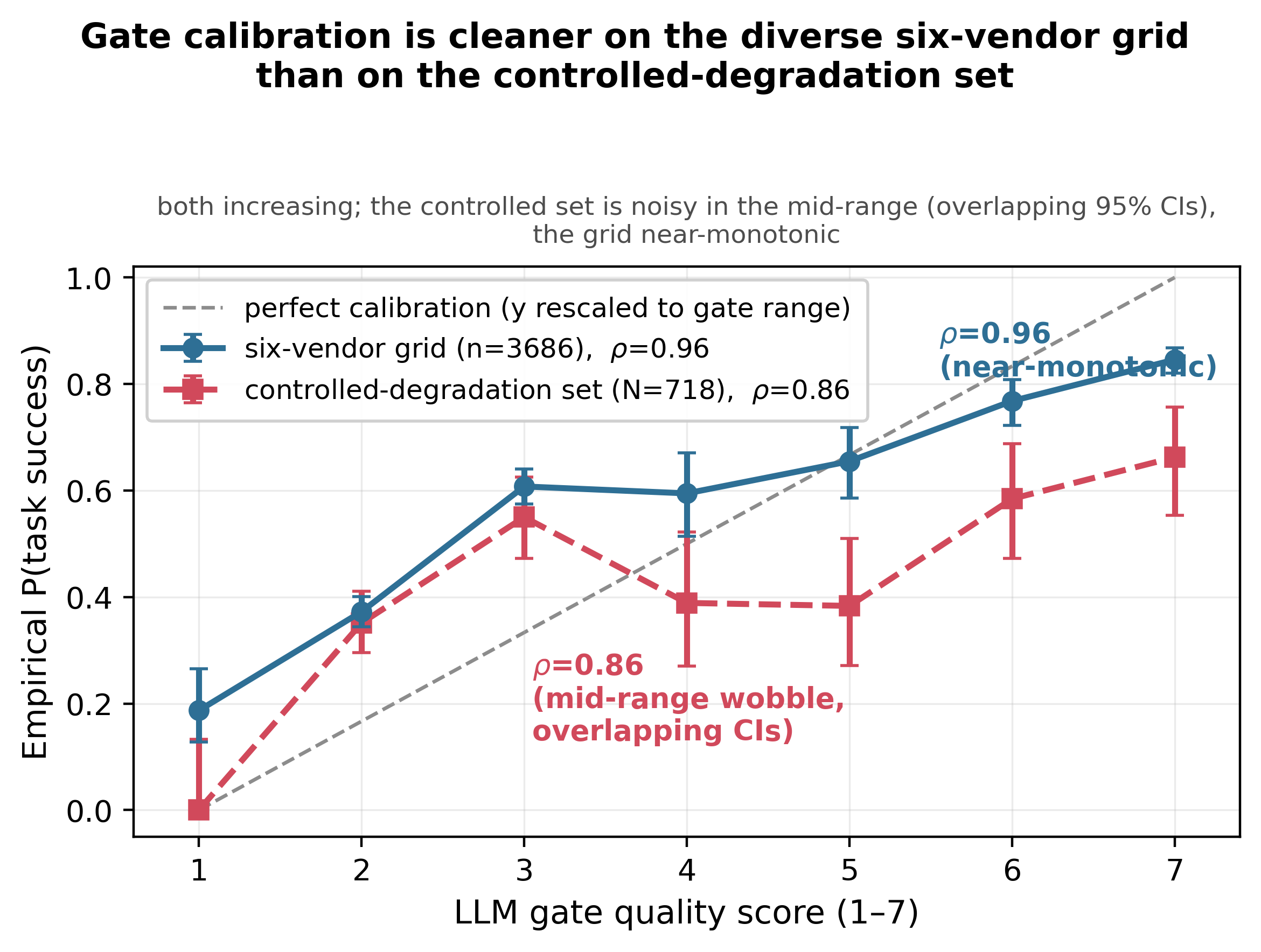}
  \caption{Gate-score calibration with Wilson 95\% confidence intervals: empirical $P(\text{success})$ at each gate
  score on the controlled-degradation set ($N{=}718$ gate-scored, Spearman \rr$=$0.86) and the six-provider grid
  ($n{=}3{,}686$ gate-scored, \rr$=$0.96). The grid is near-monotonic; the controlled set dips in the mid-range
  (scores 4--5 below score~3), but with small bins and overlapping intervals the dip is not significant.
  Calibration quality depends on the agent population.}
  \label{fig:calib}
\end{figure}

Figure~\ref{fig:calib} plots the empirical task-success rate at each integer gate score, with Wilson
95\% confidence intervals, for both the controlled-degradation set and the six-provider grid. Both are
increasing overall (Spearman \rr$=$0.86 and 0.96 over the seven score bins), so a higher gate score
does on average mean a higher chance of success. Calibration is, however, cleaner on the
diverse grid: on the controlled set the mid-range scores (4--5) sit below score~3 in point
estimate, but those bins are small ($n{=}54,60$) and their intervals overlap score~3's, so the local dip is
not individually significant. Read conservatively, the data show only that mid-range gate scores carry little
additional success signal on the degraded set; the local dip does not establish reliable
non-monotonicity. We therefore treat the gate as a \emph{ranking} instrument
(\S\ref{sec:crossfamily}) rather than a calibrated probability.

\section{Second-Provider Simulator Replication}
\label{app:secondsim}
With a controlled swap, we test directly whether the satisfied-but-failed gap is an artifact of one
over-cooperative simulator rather than a property of the satisfaction signal. Holding
\emph{everything else fixed} (the same 12 degradation variants, 6 persona strata, and identical
task ids, the same rubrics and judge, and
the same non-LLM oracle), we re-drive the grid with the user-simulator changed to a different provider,
OpenAI GPT-5.4. The simulator provider is then the only variable that differs from the matched
arm, so any persistence of the gap cannot be a single-simulator effect.

The inversion persists under the swap (Table~\ref{tab:secondsim}). Under \emph{both} simulators the
satisfied-but-failed rate sits far above the 20\% a construct-valid signal would target: GPT-5.4
66.7\% (gate) and 69.8\% (proxy) versus Sonnet-4.5 43.6\% and 52.9\% on the matched cells, all with
$p<10^{-7}$. We read this as a qualitative invariant (satisfaction fails to track success under an
independent simulator), \emph{not} as a rate comparison: the two simulators steer the same agent
into different trajectories and so yield different mean verifiable reward (GPT-5.4 0.21 vs.\ Sonnet
0.32, the harsher simulator surfacing more task failures), which mechanically shifts the absolute
false-accept rate. The defensible claim is therefore that the gap is large and highly significant
under both simulator families.

\paragraph{Full ranking grid under GPT-5.4.} The swap above isolates the \emph{construct gap}; we
also re-ran the entire 25-agent \emph{ranking} grid with the GPT-5.4 user-simulator (114 tasks per
agent, $2{,}850$ transcripts), holding agent, tasks, rubrics, judge, and oracle fixed. The agent
ordering is preserved: Spearman \rr$=$0.93 (Pearson 0.97; base-model cluster-bootstrap 95\% CI
$[0.71,0.98]$), against 0.94 under Sonnet-4.5, and the near-equal degradation reproduces (top-11
\rr$=$0.51). Same-family simulator--agent affinity therefore does not account for the ranking result
(\S\ref{sec:crossfamily}).

\begin{table}[t]
\centering\small
\setlength{\tabcolsep}{5pt}
\renewcommand{\arraystretch}{1.01}
\begin{tabular}{@{}lrrr@{}}
\toprule
User-simulator & \shortstack[r]{Mean\\rew.} & \shortstack[r]{Gate\\failed\,\%} & \shortstack[r]{Proxy\\failed\,\%} \\
\midrule
Sonnet-4.5 (Anthropic) & 0.32 & 43.6 & 52.9 \\
GPT-5.4 (OpenAI) & 0.21 & 66.7 & 69.8 \\
\bottomrule
\end{tabular}
\caption{Controlled second-provider simulator swap on the retail degradation grid: identical 12
variants $\times$ 6 strata $\times$ task ids ($n{=}216$ each), same Sonnet-4.5 agent, only the
user-simulator provider differs. The satisfied-but-failed (false-accept) rate stays far above the
20\% null under both simulators, so it is not a Sonnet-simulator artifact. ``Satisfied'' is
$\geq$5/7; ``failed'' is verifiable reward $=0$. The two simulators yield different mean agent
reward (GPT-5.4 harsher), so compare each rate to the null, not to each other.}
\label{tab:secondsim}
\end{table}

\section{Per-Agent Data Behind Figure~\ref{fig:ladder}}
\label{app:peragent}
Table~\ref{tab:peragent} gives the full per-agent values plotted in Fig.~\ref{fig:ladder} (the
cross-provider dumbbell): for each of the 25 agents, the
verifiable reward (mean\,$\pm$\,SE), the LLM-judge gate and LLM-proxy satisfaction (1--7 means),
the \emph{satisfaction surplus} (proxy mapped to $[0,1]$ minus reward), and each agent's
\textbf{rank} (1$=$best) under the verifiable reward and under the two frontier judges Opus-4.8 and
GPT-5.5. Agents are grouped by provider with providers ordered by mean reward (low$\to$high capability),
matching the figure's top-to-bottom layout. The surplus shrinks across the capability tiers
(\textbf{surplus vs.\ reward Spearman \rr$=-$0.51}, $p=8.5\mathrm{e}{-3}$, $n{=}25$): large and
positive for Mistral and Qwen ($+0.20$ to $+0.40$), mixed for Meta ($-0.06$ to $+0.20$), and near zero or negative
for the frontier (DeepSeek, OpenAI, and the strongest Anthropic configs). The rank columns make the
judges' agreements and disagreements explicit: both frontier judges place the OpenAI agents in their
top three on \emph{satisfaction}, a same-style preference the verifiable reward does not share, as it
ranks the GPT-5.4 configs at 7--9 and places Anthropic's Opus-4.6 first on ground-truth task success.

\begin{table*}[t]
\centering\small
\setlength{\tabcolsep}{6pt}
\begin{tabular}{@{}lccccccc@{}}
\toprule
& & & & & \multicolumn{3}{c}{Rank (1$=$best) by} \\
\cmidrule(l){6-8}
Agent & Reward (mean\,$\pm$\,SE) & Gate & Proxy & Surplus & Reward & Opus-4.8 & GPT-5.5 \\
\midrule
\multicolumn{8}{@{}l}{\emph{Meta} ($\bar r=0.34$)}\\
Llama-3.1-8B (t.7) & $0.32\!\pm\!0.06$ & 2.16 & 4.13 & $+0.20$ & 23 & 25 & 25 \\
Llama-3.3-70B (t.7) & $0.33\!\pm\!0.06$ & 2.32 & 2.76 & $-0.04$ & 22 & 22 & 18 \\
Llama-3.1-8B & $0.35\!\pm\!0.06$ & 2.08 & 3.79 & $+0.12$ & 19 & 24 & 24 \\
Llama-3.3-70B & $0.35\!\pm\!0.06$ & 2.18 & 2.71 & $-0.06$ & 20 & 21 & 21 \\
\midrule
\multicolumn{8}{@{}l}{\emph{Mistral} ($\bar r=0.40$)}\\
Mistral-Small & $0.28\!\pm\!0.04$ & 1.93 & 3.99 & $+0.22$ & 25 & 23 & 23 \\
Mistral-Small (t.7) & $0.30\!\pm\!0.04$ & 2.16 & 4.01 & $+0.20$ & 24 & 20 & 22 \\
Ministral-3B (t.7) & $0.35\!\pm\!0.04$ & 2.68 & 4.45 & $+0.23$ & 21 & 19 & 20 \\
Ministral-3B & $0.44\!\pm\!0.04$ & 2.69 & 4.83 & $+0.20$ & 17 & 17 & 19 \\
Mistral-Large-3 (t.7) & $0.51\!\pm\!0.04$ & 4.24 & 5.62 & $+0.26$ & 15 & 15 & 14 \\
Mistral-Large-3 & $0.54\!\pm\!0.04$ & 4.40 & 5.63 & $+0.24$ & 14 & 14 & 15 \\
\midrule
\multicolumn{8}{@{}l}{\emph{Qwen} ($\bar r=0.49$)}\\
Qwen3-32B (t.7) & $0.37\!\pm\!0.04$ & 2.54 & 5.60 & $+0.40$ & 18 & 16 & 16 \\
Qwen3-32B & $0.44\!\pm\!0.04$ & 2.66 & 5.45 & $+0.30$ & 16 & 18 & 17 \\
Qwen3-235B & $0.57\!\pm\!0.04$ & 4.16 & 5.68 & $+0.21$ & 13 & 13 & 6 \\
Qwen3-235B (t.7) & $0.59\!\pm\!0.04$ & 4.23 & 5.80 & $+0.21$ & 12 & 12 & 4 \\
\midrule
\multicolumn{8}{@{}l}{\emph{DeepSeek} ($\bar r=0.77$)}\\
DeepSeek-V3.2 (t.7) & $0.75\!\pm\!0.03$ & 5.09 & 5.53 & $+0.01$ & 8 & 9 & 5 \\
DeepSeek-V3.2 & $0.78\!\pm\!0.03$ & 5.20 & 5.62 & $-0.01$ & 6 & 5 & 7 \\
\midrule
\multicolumn{8}{@{}l}{\emph{OpenAI} ($\bar r=0.78$)}\\
GPT-5.4 (t.7) & $0.74\!\pm\!0.03$ & 5.51 & 5.57 & $+0.02$ & 9 & 3 & 3 \\
GPT-5.4 & $0.77\!\pm\!0.03$ & 5.62 & 5.56 & $-0.01$ & 7 & 2 & 2 \\
GPT-5.5 & $0.84\!\pm\!0.03$ & 5.54 & 5.79 & $-0.04$ & 3 & 1 & 1 \\
\midrule
\multicolumn{8}{@{}l}{\emph{Anthropic} ($\bar r=0.78$)}\\
Haiku-4.5 (t.7) & $0.66\!\pm\!0.04$ & 4.68 & 5.77 & $+0.13$ & 11 & 10 & 13 \\
Haiku-4.5 & $0.68\!\pm\!0.04$ & 4.71 & 5.85 & $+0.12$ & 10 & 11 & 12 \\
Sonnet-4.6 (t.7) & $0.80\!\pm\!0.03$ & 4.78 & 5.92 & $+0.02$ & 5 & 7 & 11 \\
Sonnet-4.6 & $0.81\!\pm\!0.03$ & 4.96 & 5.89 & $+0.00$ & 4 & 8 & 10 \\
Opus-4.6 & $0.84\!\pm\!0.03$ & 5.03 & 5.97 & $-0.01$ & 2 & 6 & 8 \\
Opus-4.6 (t.7) & $0.87\!\pm\!0.03$ & 5.00 & 5.97 & $-0.04$ & 1 & 4 & 9 \\
\bottomrule
\end{tabular}
\caption{Per-agent data behind Fig.~\ref{fig:ladder}, all 25 agents. Reward is the verifiable
pass-rate (mean\,$\pm$\,standard error across the agent's transcripts); Gate and Proxy are 1--7 satisfaction
means; Surplus $=$ (proxy${-}1)/6 -$ reward, i.e.\ how far satisfaction overstates verifiable
success on a common $[0,1]$ scale. The last three columns give each agent's rank (1$=$best) by
verifiable reward and by the two frontier judges Opus-4.8 and GPT-5.5. Providers are ordered by mean
reward $\bar r$. Surplus vs.\ reward: Spearman \rr$=-$0.51 ($p=8.5\mathrm{e}{-3}$). The rank columns
expose where judges depart from ground truth: both frontier judges rank the GPT agents 1--3 on
satisfaction, well above their reward ranks (GPT-5.4 at 7--9), a same-style preference, whereas the
verifiable reward ranks Anthropic's Opus-4.6 first; the judges still track the overall ordering
(full-ladder \rr$\geq$0.84, \S\ref{sec:crossprovider}).}
\label{tab:peragent}
\end{table*}

\section{Per-Judge Ranking Validity Across Providers and Domains}
\label{app:perjudge}
\begin{figure*}[t]
  \centering
  \includegraphics[width=\textwidth]{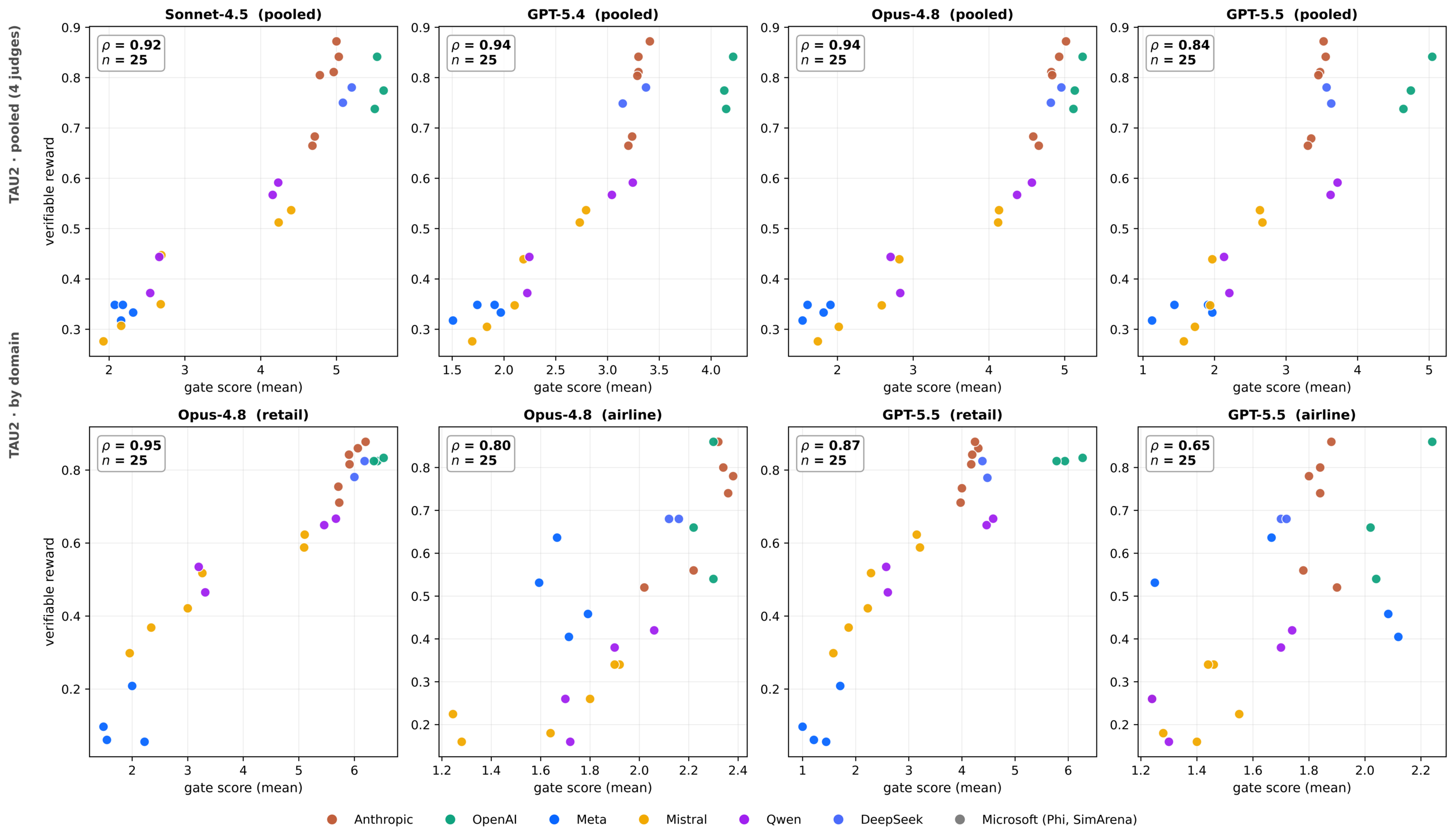}
  \caption{Per-judge gate-vs-reward validity across all available judges and domains. $\tau^2$ rows:
  four judges pooled (Sonnet-4.5, GPT-5.4, Opus-4.8, GPT-5.5) and the Opus-4.8/GPT-5.5 retail vs
  airline splits; each point is an agent, provider-colored, with Spearman \rr\ annotated.}
  \label{fig:appendix3c}
\end{figure*}

Section~\ref{sec:crossfamily} summarizes ranking validity; Fig.~\ref{fig:appendix3c} shows the
underlying per-judge gate-vs-reward scatters in full. On the $\tau^2$ grid the gate's ranking is
recovered by all four judges, the primary Opus-4.8 (\rr$=$0.94) and GPT-5.5 (0.84),
and the deployed Sonnet-4.5 (0.92) and GPT-5.4 (0.94); the two frontier judges' domain splits are
Opus-4.8 retail 0.95, airline 0.80, and GPT-5.5 retail 0.87, airline 0.65.

\subsection{The near-equal regime persists under frontier judges}
\label{app:frontier}
\begin{figure}[t]
  \centering
  \includegraphics[width=\columnwidth]{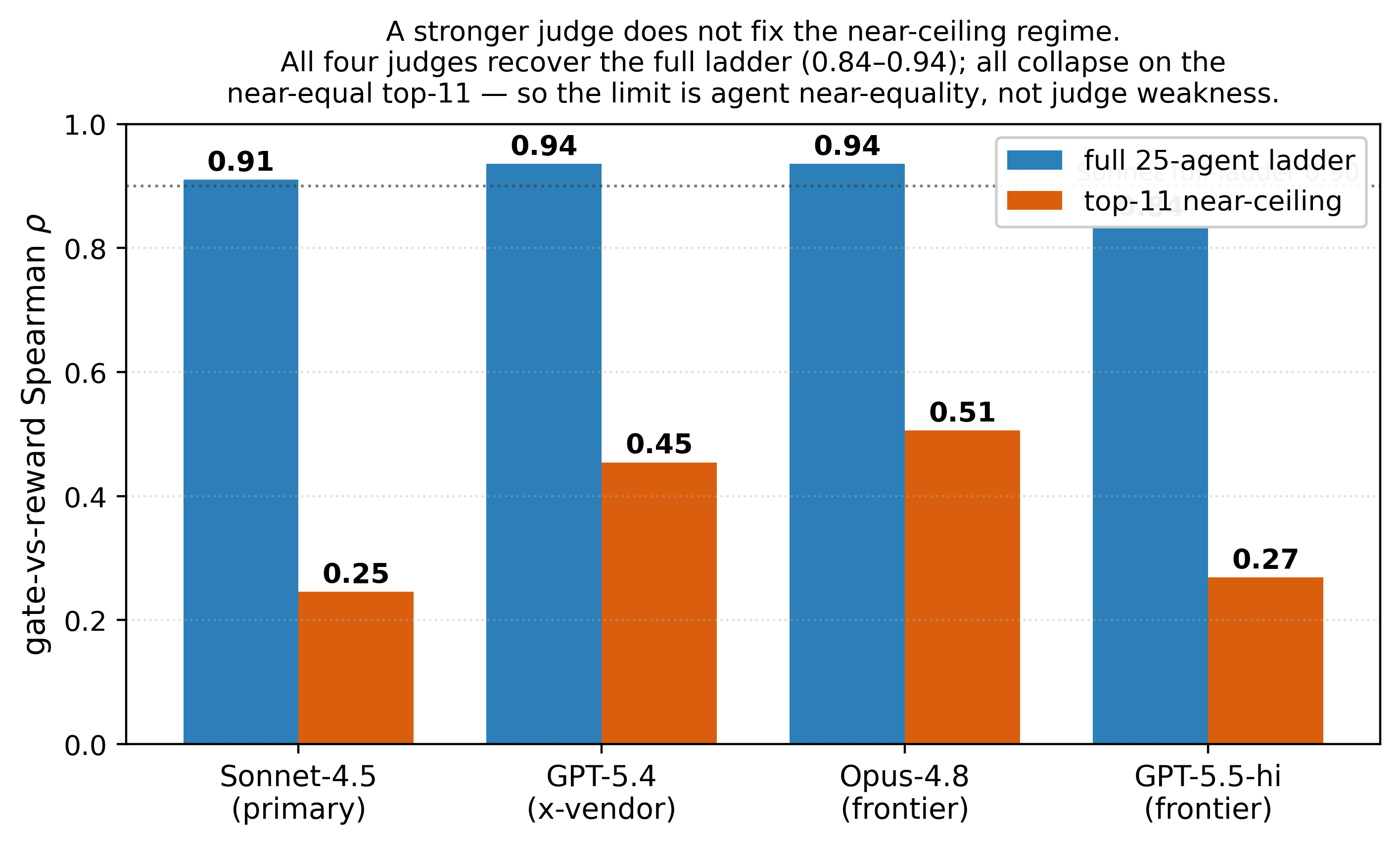}
  \caption{Four-judge robustness. All four judges, including the frontier Opus-4.8 and GPT-5.5,
  which match the strongest agents in the pool by verifiable pass-rate, recover the full
  25-agent ladder (\rr$=$0.84--0.94), but none recovers the near-equal top-11 (\rr$=$0.25--0.51, all
  n.s.). The near-ceiling regime persists even for the most capable judges, so the limit reflects
  agent near-equality bounded by oracle resolution, not judge capability.}
  \label{fig:frontierjudges}
\end{figure}

We test whether the judge is weaker than the strongest agents it scores, which would make the near-equal
top-11 ceiling (\S\ref{sec:crossfamily}) an artifact of judge capability rather than a real
property of the agents. Figure~\ref{fig:frontierjudges} rules this out. It places, for all four
judges, the gate-vs-reward Spearman \rr\ on the full 25-agent ladder beside the same \rr\ on the
near-equal top-11 subset. Two patterns stand out: every judge recovers the full
ladder (\rr$=$0.84--0.94), and on the top-11 no judge recovers the ordering
(\rr$=$0.25--0.51, all n.s. at $n{=}11$). The two frontier judges (Opus-4.8, GPT-5.5),
which match the strongest agents in the pool by verifiable pass-rate (\S\ref{sec:crossprovider}),
do no better there than the deployed Sonnet-4.5 judge. If the ceiling were a judge-capability artifact, the
most capable judges would escape it. The near-equal limit therefore reflects
agent near-equality bounded by oracle resolution, not judge capability.

\paragraph{Decision-disagreement rate (\S\ref{sec:crossfamily}).} Table~\ref{tab:flips} gives, per
signal, the fraction of agent pairs on which the gate promotes the lower-reward agent, split by
whether the pair's verifiable rewards are near-equal ($|\Delta R|<0.1$, $n{=}87$ pairs) or wide
($|\Delta R|\geq0.1$, $n{=}211$). Every signal is near-perfect on wide pairs and degrades sharply on
close ones; the effect is not sensitive to the exact 0.1 threshold (Opus-4.8: 45\%, 31\%, 26\% at
$|\Delta R|<0.05/0.10/0.15$).

\begin{table}[t]
\centering\small
\setlength{\tabcolsep}{6pt}
\renewcommand{\arraystretch}{1.05}
\begin{tabular}{@{}lcc@{}}
\toprule
Signal & Near-equal\,\% & Wide\,\% \\
\midrule
Opus-4.8 gate (primary frontier) & 31.0 & 0.9 \\
GPT-5.4 gate & 29.9 & 1.4 \\
GPT-5.5 gate & 39.1 & 7.6 \\
Pooled gate (\texttt{gate\_quality}) & 37.9 & 1.4 \\
LLM-proxy satisfaction & 29.9 & 10.4 \\
\bottomrule
\end{tabular}
\caption{Decision-disagreement rate by signal: fraction of pairs where the gate promotes the
\emph{lower}-reward agent, on near-equal ($|\Delta R|<0.1$, $n{=}87$) vs.\ wide ($|\Delta R|\geq0.1$,
$n{=}211$) pairs; the 2 exact-tie pairs of the $\binom{25}{2}{=}300$ are excluded. Base-model
cluster-bootstrap 95\% CI on the Opus-4.8 near-equal rate: $[11.6, 50.0]$.}
\label{tab:flips}
\end{table}

\paragraph{Is the 31\% reference-ranking noise or an artifact of non-independent pairs?} Two checks
rule this out. A base-model cluster bootstrap (13 clusters, resampling base models rather than pairs) puts
the primary Opus-4.8 near-equal rate at 31.0\%, 95\% CI $[11.6, 50.0]$, its lower bound still
${\sim}13{\times}$ the 0.9\% wide-pair rate; and dropping the least-independent same-base-model
temperature pairs leaves 32.0\% on the 75 cross-base-model pairs (24/75). The disagreement is also
signal-\emph{specific} rather than shared ordering noise: of the 60 near-equal pairs flipped by at
least one of the five signals, only 4 are flipped by all five and 22 by exactly one (mean pairwise
Jaccard 0.40), so the 31\% is a population-level, signal-specific property, not shared
reference-ranking noise.

\subsection{Same-family self-preference}
\label{app:selfpref}
\begin{figure}[t]
  \centering
  \includegraphics[width=0.92\columnwidth]{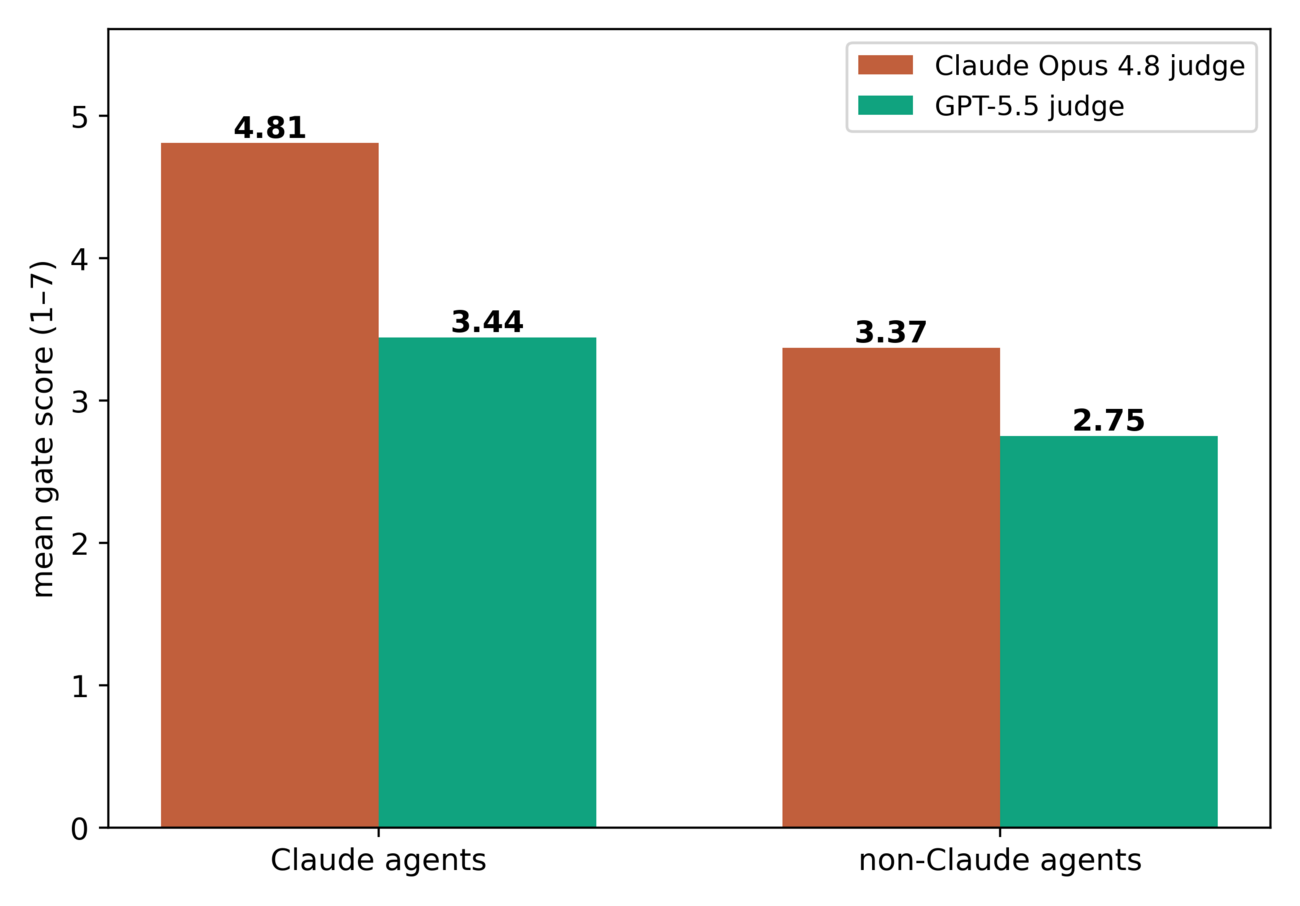}
  \caption{Same-family self-preference, shown explicitly. The Opus-4.8 judge inflates Claude
  agents relative to the GPT-5.5 judge more than it inflates non-Claude agents
  (diff-in-diff $+0.75$/7).}
  \label{fig:selfpref}
\end{figure}

The two-provider design isolates same-family self-preference as a difference of differences
(\S\ref{sec:crossprovider}): the Anthropic Opus-4.8 judge rates Claude agents above the
GPT-5.5 judge, and by a \emph{larger} margin than it rates non-Claude agents
(Fig.~\ref{fig:selfpref}). The diff-in-diff is $+0.75$/7 (the deployed Sonnet-4.5 judge shows the
same effect at $+0.67$/7 against GPT-5.4); being scale-invariant, it reflects
same-family self-preference rather than absolute harshness. Crucially, the \emph{ranking} agreement is
unchanged (\rr$=$0.92; \S\ref{sec:crossprovider}), so self-preference is a measured caveat
on the absolute scores while ranking validity holds.

\subsection{Judge stability}
\label{app:judgestability}
Two reruns confirm the agent ranking is not an artifact of judge noise. First, in a test--retest,
re-scoring 48 transcripts three times (judge temperature 0.7) yields a judge self-reliability of
ICC$=$0.87 \citep{shrout1979intraclass} and a variant-level rank stability of \rr$=$0.92 across re-scorings, a judge-noise
measure distinct from the human-panel inter-annotator agreement (Krippendorff's $\alpha{=}0.79$).
Second, running the GPT judge five independent times agrees with itself on the agent
ranking at mean \rr$=$0.74 (min 0.52); per-run validity against the verifiable reward is noisier
(\rr$=$0.24--0.59 across runs), which is why we report that judge as a ranking-agreement
check (\S\ref{sec:crossprovider}) rather than a per-run validity estimate.

\section{Cheap Signals for the Near-Equal Regime}
\label{app:cheap}
We test whether any cheaper signal removes the near-equal decision error of \S\ref{sec:crossfamily}
without the paid audit. Across 21 candidate signals, spanning single judges, the process-blind proxy,
the completion bit, and unweighted and confidence-weighted judge ensembles, none outperforms the best single
judge on near-equal pairs. Averaging all four judges gives exactly 31.0\% (27/87), identical to the
best single judge, because the judges are highly correlated (pairwise \rr\ 0.67--0.98) and flip the
same close pairs: the error is \emph{common-mode}, so ensembling cannot average it away. The one
effective lever is \emph{calibrated abstention}, declining to rank a pair whose gate-score gap falls inside
its bootstrap sampling noise. On the pairs it does rank (31--49\% of near-equal pairs, depending on the
abstention threshold) it roughly halves the error to 14.8\%, buying selectivity rather than
resolution, which confirms that the limit is oracle resolution, not judge choice.

\section{Why Airline Ranking Is Weaker: Gate-Score Range Restriction}
\label{app:rangerestriction}
\begin{figure*}[t]
  \centering
  \includegraphics[width=\textwidth]{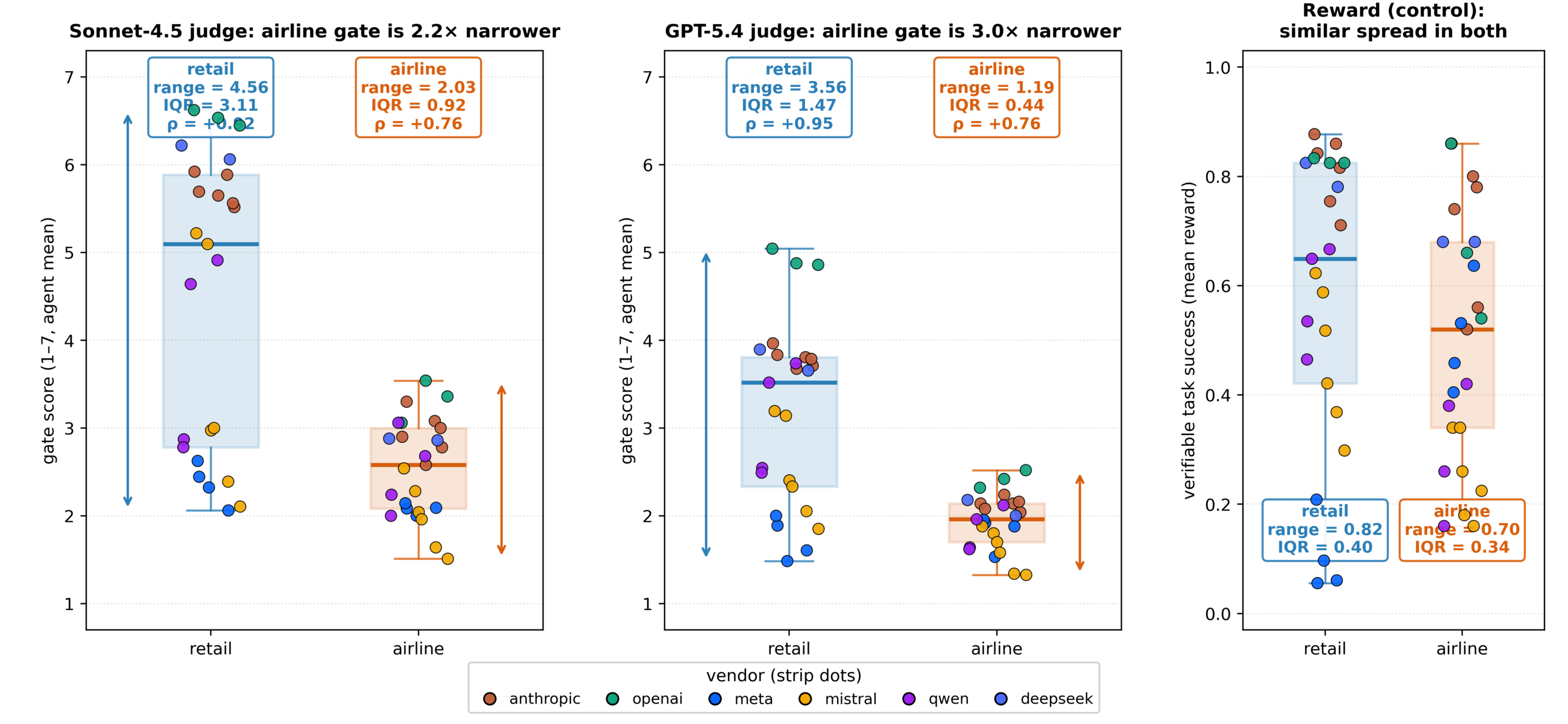}
  \caption{Gate-score distributions by domain for both judges (box $+$ per-agent strip), with the
  reward distribution as a control. The airline gate signal is $\approx$2.2--3.0$\times$ more
  compressed than retail (Sonnet and GPT-5.4 alike), whereas verifiable reward is comparably spread
  in both domains, so the weaker airline ranking \rr\ is driven by gate-score range restriction,
  not by the gate being less valid there.}
  \label{fig:appendix6b}
\end{figure*}

\emph{Range restriction}, rather than a deficiency of the gate, explains the lower airline ranking
validity (\S\ref{sec:crossfamily}). Fig.~\ref{fig:appendix6b} shows the gate-score distributions per domain:
the airline gate spans a $\approx$2.2$\times$ narrower band than retail (Sonnet gate range 2.03 vs
4.56; GPT-5.4 gate 1.19 vs 3.56, a 3.0$\times$ compression), while the verifiable-reward spread is
nearly identical across domains (range ratio 1.17$\times$). With agents compressed into a narrow gate
band there is less to rank, which attenuates a rank correlation; both judges exhibit the same
compression, confirming it is a property of the airline substrate rather than of either judge.

\section{Persona Strata}
\label{app:personas}
The user-simulator is conditioned on one of six persona strata, each a 2--4 sentence second-person
behavioral overlay prepended to the $\tau^2$-bench task instructions (the task facts, namely order
IDs, customer identity, and item specs, are held fixed, so only tone, cooperativeness, and pacing
change; the verifiable-reward signal is unaffected). The strata are hand-authored archetypes whose
axes (cooperativeness, patience, assertiveness) are informed by the OPeRA persona schema
\citep{opera2026} (demographics, Big-Five, and Consumer-Styles-Inventory shopping styles); we do not
sample individual OPeRA records. Table~\ref{tab:personas} lists them. Persona conditioning is
behaviorally real: on an identical retail task, the cooperative user writes ``\emph{Yes, I confirm.
Please go ahead\,\ldots\ I appreciate your help}'' while the impatient user writes ``\emph{I don't
have time for this\,\ldots\ just do it already}'' and the skeptical negotiator probes ``\emph{this
`no modification fee' sounds too good to be true, what's the catch?}''

\begin{table}[t]
\centering\small
\setlength{\tabcolsep}{4pt}
\begin{tabular}{@{}llll@{}}
\toprule
Stratum & Coop. & Patience & Assert. \\
\midrule
S1 Cooperative & high & high & low \\
S2 Impatient/rude & low & low & high \\
S3 Distracted & med & med & low \\
S4 Anxious/low-trust & med & med & med \\
S5 Terse & high & med & med \\
S8 Skeptical negotiator & low & med & high \\
\bottomrule
\end{tabular}
\caption{The six persona strata and their design axes (cooperativeness, patience, assertiveness),
informed by the OPeRA schema \citep{opera2026}. The six are drawn from an eight-archetype pool;
S6 and S7 (chatty over-sharer, indecisive perfectionist) are defined but not exercised in the
reported runs, hence the index gap. Used in the controlled-degradation set and the
150-transcript human panel; see \S\ref{sec:personas} for their effect on agent difficulty and
ranking.}
\label{tab:personas}
\end{table}

\begin{figure}[t]
  \centering
  \includegraphics[width=\columnwidth]{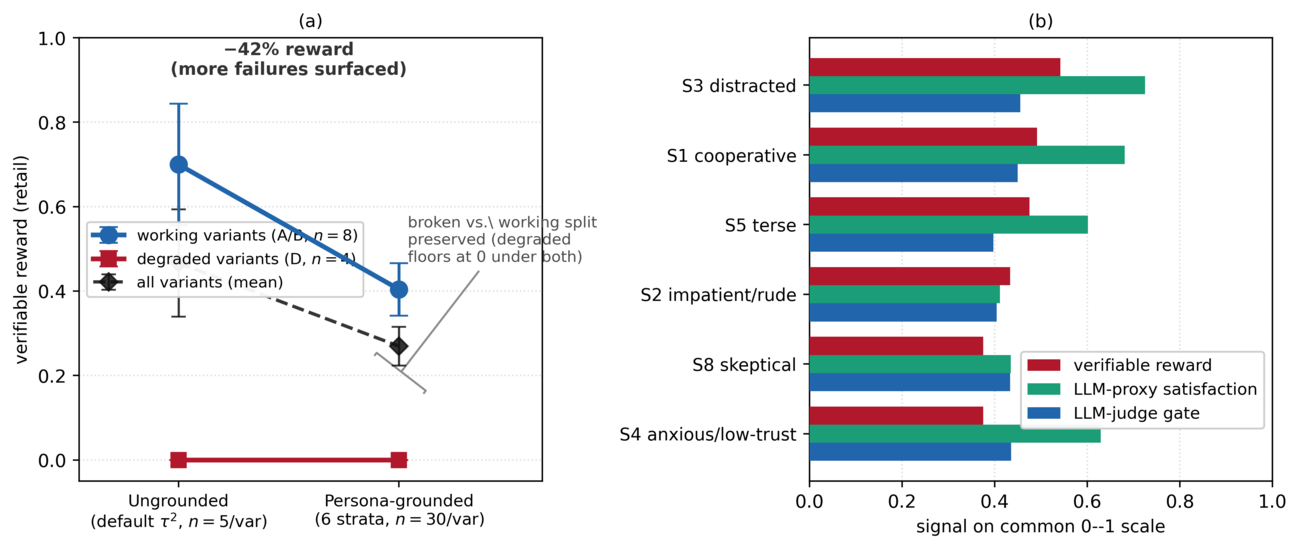}
  \caption{Effect of persona strata on agent difficulty and ranking (retail), discussed in
  \S\ref{sec:personas}. \textbf{(a)} Per-variant verifiable reward under the ungrounded
  default-$\tau^2$ user versus the six persona strata: personas cut mean verifiable reward $42\%$
  ($0.47\to0.27$) while the agent ordering holds (reward-rank \rr$=$0.83). \textbf{(b)}
  Per-stratum reward, gate, and LLM-proxy satisfaction (grounded $720$-set, common $[0,1]$ scale):
  satisfaction varies sharply by persona but does not track reward.}
  \label{fig:persona}
\end{figure}

\paragraph{Verbatim persona overlays.} Each stratum is a second-person behavioral overlay prepended
to the task instructions with the wrapper ``\texttt{PERSONA: adopt this style for the entire
conversation; keep all facts of your request unchanged: \{overlay\}}'', followed by the unchanged
$\tau^2$-bench task. The full overlays are:
\begin{description}\setlength{\itemsep}{2pt}
\item[S1 Cooperative.] A courteous, even-tempered customer in their mid-30s who shops online
regularly and trusts the representative. Answers directly and completely, stays focused, stays
patient through multi-step resolution, confirms details without rushing, and expresses mild
appreciation when things go well.
\item[S2 Impatient/rude.] A busy, short-tempered customer who feels they have wasted too much time.
Blunt to the point of rudeness, interrupts with demands rather than answering procedural questions,
becomes openly irritated at requests for information they think the agent should already have,
pressures the agent to skip steps, and demands escalation when anything takes more than a couple of
exchanges.
\item[S3 Distracted.] A multitasking customer only half paying attention, replying while distracted.
Gives incomplete or slightly-off answers, misses what was just asked, and drifts onto tangents
before returning; not hostile, just scattered (``sorry, what did you ask?''), so the agent must keep
them on track.
\item[S4 Anxious/low-trust.] A risk-averse customer who distrusts remote support and worries
something will go wrong. Asks the agent to justify each action before agreeing, seeks repeated
reassurance about charges, and hesitates before sharing details; polite but cautious (``are you
sure?'', ``what happens if this doesn't work?'').
\item[S5 Terse.] An experienced, no-nonsense customer who values speed. Responds in clipped, minimal
phrases, skips pleasantries, and provides only the specific information requested; cooperative in
substance but cold in style, nudging the agent to get to the point.
\item[S8 Skeptical negotiator.] A price-sensitive, deal-driven customer who treats every interaction
as a negotiation and is skeptical of any policy that does not favor them. Assertively challenges
fees and refusals, presses for waivers or exceptions, and probes for loopholes; persistent and
adversarial but not abusive, relenting only once a limit is firmly justified.
\end{description}
The persona modifies only tone, cooperativeness, pacing, and assertiveness; task facts
(\texttt{reason\_for\_call}, \texttt{known\_info}, \texttt{evaluation\_criteria}) are untouched, so
the verifiable reward is unaffected. The two defined-but-unexercised strata (S6 chatty over-sharer,
S7 indecisive perfectionist) account for the gap between S5 and S8.

\section{Dataset Statistics}
\label{app:datastats}
Table~\ref{tab:datastats} summarizes the three evaluation substrates. The cross-provider ladder is
the headline grid; the controlled-degradation set is the positive control (\S\ref{sec:freebaseline});
the human panel and SimulatorArena provide the human-grounded checks (\S\ref{sec:satnotsuccess}).
The controlled-degradation set and the persona-simulated conversations are generated synthetically;
synthetic generation is a common remedy for scarce labeled data in NLP \citep{selfinstruct2023,lentex2025}.
The ladder's 14 base models are Anthropic Opus-4.6/Sonnet-4.6/Haiku-4.5 \citep{claudeopus46card,claudesonnet46card,claudehaiku45card}, Meta Llama 405B/70B/8B
\citep{grattafiori2024llama3herdmodels,llama33modelcard}, Mistral Large-3/Small/Ministral-3B \citep{mistral3}, Qwen3 235B/32B \citep{yang2025qwen3technicalreport},
DeepSeek-V3.2 \citep{deepseekv3,deepseekv32modelcard}, and OpenAI GPT-5.4/5.5 \citep{openai2026gpt54,openai2026gpt55}, each run at two temperatures.

\begin{table}[t]
\centering\small
\setlength{\tabcolsep}{4pt}
\renewcommand{\arraystretch}{1.05}
\begin{tabular}{@{}lr@{}}
\toprule
\multicolumn{2}{@{}l}{\emph{Cross-provider ladder ($\tau^2$-bench retail + airline)}}\\
Scorable transcripts (non-null reward) & 3{,}691 \\
\quad retail / airline & 2{,}487 / 1{,}204 \\
Agent configurations scored & 25 \\
Task pool (retail / airline) & 114 / 50 \\
\midrule
\multicolumn{2}{@{}l}{\emph{Controlled-degradation set (Sonnet-4.5, flag-only)}}\\
Transcripts & 720 \\
Cells ($12$ configs $\times\,6$ strata $\times\,2$ domains) & 144 \\
Transcripts per cell & 5 \\
Tiers (good / medium / degraded) & 3 / 5 / 4 configs \\
\midrule
\multicolumn{2}{@{}l}{\emph{Human panel ($\tau^2$ subsample)}}\\
Transcripts / annotators & 150 / 3 \\
Stratification (retail / airline) & 85 / 65 \\
\quad failed / succeeded & 86 / 64 \\
Krippendorff's $\alpha$ (satisfaction) & 0.79 \\
\midrule
\multicolumn{2}{@{}l}{\emph{SimulatorArena (math tutoring)}}\\
Models / graded conversations & 9 / 50 \\
High-satisfaction subset ($\geq$8/10) & 31 \\
\bottomrule
\end{tabular}
\caption{Dataset statistics across the three substrates. The 25 scored agent configurations are the
28 (model, temperature) cells minus three excluded for infrastructure reasons (Llama-3.1-405B at
both temperatures, throttle-limited; and the GPT-5.5 temperature-0.7 cell, as GPT-5.5 fixes its
sampling setting). Conversations average 31 messages, 8.6 tool calls, and 7.6 user turns.}
\label{tab:datastats}
\end{table}

\section{Oracle Patch and Validation}
\label{app:oracle}
The verifiable reward is the ground truth the audit rests on, so we document our only modification to
the $\tau^2$-bench evaluator, a bug-fix rather than a degree of freedom, in two disjoint parts.

\paragraph{(A) Backend swap (no scoring change).} Four \texttt{DEFAULT\_LLM\_*} constants are
repointed from \texttt{gpt-4.1} to \texttt{bedrock/\dots claude-sonnet-4-5} (temperatures stay at 0),
so the agent, simulator, and LLM sub-evaluator run on Bedrock; no scoring logic is touched.

\paragraph{(B) Tolerant verdict parser.} A \texttt{loads\_tolerant} helper replaces a bare
\texttt{json.loads} at two sites (tool-output comparison and NL-assertion parsing): it tries strict
\texttt{json.loads} \emph{first} and only on failure strips \texttt{```json} fences or extracts the
outermost JSON object, re-raising if nothing parses. It never rewrites or flips a parsed verdict and
adds no pass condition; it only recovers verdicts strict parsing would drop because Claude wraps JSON
in prose (absent in GPT JSON-mode). A 16-case unit test confirms identical parsing on valid JSON,
recovery on fenced/wrapped JSON, and a raise on unparseable input.

\paragraph{Why the reward is robust to the patch.} The reward is the product of each task's
\texttt{reward\_basis} components; the DB-state, action, environment-assertion, and \texttt{communicate}
checks are deterministic Python, and the NL-assertion judge is the only LLM the parser touches. Airline
(\texttt{reward\_basis}\,$=$\,(DB, COMMUNICATE)) is thus fully deterministic and patch-independent; for
retail ((DB, NL\_ASSERTION)) the NL term is multiplicatively gated by the deterministic DB check and
changed the outcome in only 113 of 2{,}537 applicable simulations (4.5\%). The harness persists only the
parsed verdict, so we rely on this lossless-by-construction argument rather than an empirical recovery
rate, and make no claim of human--oracle agreement (the panel rated experience, not task success).
Crucially, the satisfaction--success gap does not depend on this component: on \emph{airline}, whose
oracle contains no LLM, the process-blind proxy still false-accepts 62.1\% of satisfied conversations
($325/523$).

\section{Prompts and Rubrics}
\label{app:prompts}
We reproduce the operative prompts verbatim. Model identifiers
and tool names are kept as run. All scoring runs offline through a Bedrock-Converse harness; a
tolerant judge-output parser recovers the score line from formatting variation, cutting null judge
scores from 36.6\% to near zero.

\paragraph{Release-gate judge (policy-aware).} System prompt (abridged): \emph{``You are a release-gate
quality auditor for an automated retail customer-service agent\ldots\ you evaluate ONLY the assistant,
from the standpoint of an operations supervisor deciding whether this agent variant is safe to
ship.''} It scores four dimensions in priority order, (1)~policy adherence (authenticate before
acting; explicit confirmation before any database write; one tool call per turn; no fabrication),
(2)~task resolution, (3)~efficiency, and (4)~professionalism, on a 1--7 scale, explicitly instructing
that ``a polite agent that skipped authentication or wrote to the database without confirmation must
score low,'' and emitting a final \texttt{SCORE: <1--7>} line.

\paragraph{Human-proxy (satisfaction-only).} System prompt (abridged): \emph{``You are role-playing a
real online shopper who just finished a chat\ldots\ answer the way an ordinary, non-expert customer
would, based purely on how the conversation FELT to you, not on any knowledge of company policies,
internal tools, or what the agent did behind the scenes.''} It rates satisfaction, respect/tone,
clarity, perceived helpfulness, and would-return on 1--7, and is shown only the user-visible spoken
turns (tool calls and the task stripped out), enforcing process-blindness. This disjointness in role,
evidence, and vocabulary is what prevents the gate and proxy from agreeing by shared-rubric artifact.

\paragraph{Human panel.} The three annotators received written instructions to rate \emph{the feeling
of being served, not the mechanics}: ``An agent can feel great yet not actually solve your problem,
and vice versa; rate your experience.'' They scored the same five dimensions (1--7) plus a one-phrase
``biggest problem'' field, blind to the verifiable reward, variant, and model identity, on the
user-visible transcript only.

\paragraph{User simulator.} The persona-driven simulator extends the base $\tau^2$-bench guidelines
(``generate one message at a time\ldots\ disclose information progressively\ldots\ never hallucinate
information not in the scenario\ldots\ emit \texttt{\#\#\#STOP\#\#\#} when the goal is satisfied'') with
the persona overlay (Appendix~\ref{app:personas}); task facts are unchanged.

\section{Representative Examples}
\label{app:examples}
One real, de-identified example per substrate (public-benchmark tasks; no customer data).

\paragraph{$\tau^2$ retail (satisfied but failed).} Task: a customer (Yusuf Rossi, zip 19122)
asks how many t-shirt options exist and to change all pending t-shirt orders to purple, size~S,
v-neck, polyester. The agent authenticates by name+zip, lists the options, and reports
``\emph{Both modifications have been completed successfully}.'' The user ends satisfied
(\texttt{\#\#\#STOP\#\#\#}). \textbf{Gate 7/7, proxy 7/7, human panel 5.5/7, verifiable reward 0.0}: all three satisfaction
signals passed, yet a pending order was never modified (12 of 13 tool calls correct), so the task
failed despite a flawless-feeling interaction.

\paragraph{$\tau^2$ airline (satisfied but failed).} Task: a customer (Mohamed Silva) asks for the
summed gift-card and certificate balances and to rebook a reservation to the cheapest business
round-trip without changing dates. The agent reports balances and confirms the charges; the user
replies ``\emph{That's everything I needed. Thank you so much!}'' \textbf{Proxy 7/7, reward 0.0}: the
rebooking did not match the required end-state, and the cross-provider judge (GPT-5.4 gate 1/7) flags
it.

\paragraph{SimulatorArena math tutoring (satisfied but wrong).} Problem: 11 players each pass to
every other player three times; how many passes? The tutor walks through ``one player makes
$10\times3=30$ passes,'' then states ``the correct total is indeed 165. Well done!'' \textbf{Human
rating 10/10, proxy 7/7, but verifiably incorrect}: the tutor's own shown work ($30$ passes per player, $11$ players) implies $330$,
not the stated $165$; the policy-aware gate (Opus-4.8 1/7) catches the error the human did
not.

\end{document}